%% file: 0_main.tex
\documentclass{article}

\usepackage[preprint]{neurips_2023}

\usepackage[utf8]{inputenc} 
\usepackage[T1]{fontenc}    
\usepackage{hyperref}       
\usepackage{url}            
\usepackage{booktabs}       
\usepackage{amsfonts}       
\usepackage{nicefrac}       
\usepackage{microtype}      
\usepackage[table,xcdraw]{xcolor}         

\usepackage{subcaption}
\usepackage{wrapfig}
\usepackage{multirow}
\usepackage{makecell}
\usepackage{xspace}
\usepackage{enumitem}
\usepackage{amsmath}
\usepackage{listings}
\usepackage{algorithm}
\usepackage{algorithmic}
\usepackage{amsfonts,amssymb}
\usepackage{mathrsfs}
\usepackage{subcaption}
\usepackage{natbib}
\setcitestyle{numbers,square}
\usepackage{graphicx}       
\usepackage{cleveref}
\usepackage{fontawesome}
\usepackage{utfsym}
\usepackage[most]{tcolorbox}

\usepackage{tabularx}
\usepackage{CJKutf8}

\newcommand{\hhide}[1]{}
\newcommand{\hide}[1]{}

\newcommand{\aspace}{\hspace{1em}}

\newtcolorbox{promptbox}[1][]{
  breakable,
  title=#1,
  colback=gray!5,
  colframe=black,
  colbacktitle=gray!15,
  coltitle=black,
  fonttitle=\bfseries,
  bottomrule=1.5pt,
  toprule=1.5pt,
  leftrule=1pt,
  rightrule=1pt,
  arc=0pt,
  outer arc=0pt,
  enhanced,
  before upper={\parindent=1.5em} 
}

\title{CogVLM Family: Technical Report}
\title{CogVLM2: Visual Language Models for Image and Video Understanding}

 \author{
Wenyi Hong\aspace  Weihan Wang\aspace {\bf Ming Ding}\aspace  Wenmeng Yu\aspace Qingsong Lv\aspace Yan Wang\aspace     \\
{\bf Yean Cheng}\aspace {\bf Shiyu Huang}\aspace {\bf Junhui Ji}\aspace {\bf Zhao Xue}\aspace  {\bf Lei Zhao}\aspace  {\bf Zhuoyi Yang}\aspace  {\bf Xiaotao Gu}\aspace  \\
{\bf Xiaohan Zhang}\aspace  {\bf Guanyu Feng}\aspace  {\bf Da Yin}\aspace  {\bf Zihan Wang}\aspace  {\bf Ji Qi}\aspace {\bf Xixuan Song}\aspace \\
{\bf Peng Zhang}\aspace {\bf Debing Liu}\aspace  {\bf Bin Xu}\aspace  {\bf Juanzi Li}\aspace  {\bf Yuxiao Dong}\aspace {\bf Jie Tang}
~\\
$^{1}$Zhipu AI \aspace $^{2}$Tsinghua University\\
 \\
{\includegraphics[height=3.5ex]{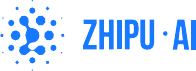}}
 }

\begin{document}

\maketitle
\renewcommand{\thefootnote}{}
\footnotetext{Core contributors: Wenyi, Weihan, Ming, Wenmeng, Qingsong, Yan, Yean, Shiyu and Junhui.}
\renewcommand{\thefootnote}{\arabic{footnote}}

\begin{abstract}

Beginning with VisualGLM and CogVLM, we are continuously exploring VLMs in pursuit of enhanced vision-language fusion, efficient higher-resolution architecture, and broader modalities and applications. 
Here we propose the CogVLM2 family, a new generation of visual language models for image and video understanding including CogVLM2, CogVLM2-Video and GLM-4V. 
As an image understanding model, CogVLM2 inherits the visual expert architecture with improved training recipes in both pre-training and post-training stages, supporting input resolution up to $1344 \times 1344$ pixels. As a video understanding model, CogVLM2-Video integrates multi-frame input with timestamps and proposes automated temporal grounding data construction.
Notably, CogVLM2 family has achieved state-of-the-art results on benchmarks like MMBench, MM-Vet, TextVQA, MVBench and VCGBench. 
All models are open-sourced in \url{https://github.com/THUDM/CogVLM2} and \url{https://github.com/THUDM/GLM-4}, contributing to the advancement of the field.

\end{abstract}

\input{1_intro}

\input{2_models}
\input{3_training}
\input{4_evaluation}

\input{5_discussion}

\clearpage

\bibliographystyle{abbrv}
\bibliography{ref}

\clearpage

\appendix

\input{appendix/TQA_details}
\input{appendix/CogVLM2_cases}
\input{appendix/mvbench}

\end{document}

%% file: 1_intro.tex
\begin{figure}[h]
\begin{center}
\includegraphics[width=\linewidth]{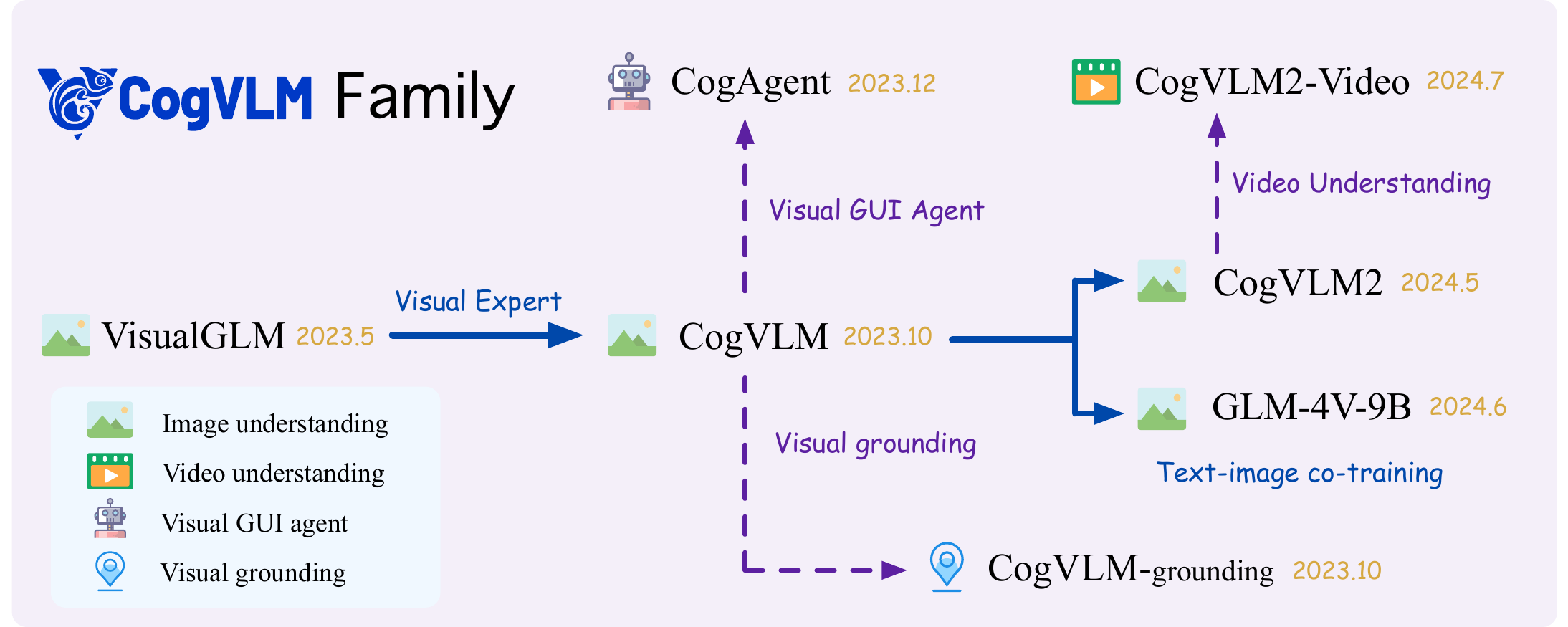}
\end{center}
\caption{Overview of CogVLM family.}
\label{fig:cogvlm_roadmap}
\end{figure}

\section{Introduction}




In recent years, large language models (LLMs) have demonstrated increasingly powerful language comprehension and generation capabilities, and are gradually helpful to our daily life. Despite their extensive knowledge and strong capability, LLMs are limited to accepting only text-based input, which significantly restricts their range of applications and prevents them from acquiring broader knowledge from the vast amounts of visual data available. Therefore, how to equip LLMs with the capability of perceiving the world visually, i.e. to develop visual language models (VLMs) has become a popular research topic. 

Since April 2023, CogVLM Team are actively engaged in the research of visual language models.
As shown in \Cref{fig:cogvlm_roadmap}, we have successively \textbf{open-sourced} a family of visual language models covering a wide range of modalities and capabilities including image understanding, video understanding, visual grounding, GUI (Graphical User Interfaces) agent, etc. 

As our initial effort in developing a visual language model, \textbf{VisualGLM}
is the first open-source multi-modal dialog language model that supports both Chinese and English. The model is built on an architecture similar to BLIP-2~\cite{li2023blip} and is trained based on the ChatGLM-6B.

While VLMs are all built upon powerful LLMs, we have observed that VLMs typically employ shallow alignment techniques to connect image features with the LLM, which hinders deep integration and understanding between visual and linguistic modalities. To overcome this, we proposed \textbf{CogVLM}~\cite{wang2023cogvlm}
and its grounding version \textbf{CogVLM-grounding}. CogVLM features the design of visual experts, which enables a deep fusion of vision language features without sacrificing natural language performance. 

To explore VLM's application in the domain of GUI agents, we proposed \textbf{CogAgent}~\cite{hong2024cogagent}
based on CogVLM's robust visual understanding capabilities. We propose a high-resolution cross-module to efficiently enhance the image resolution processed by the model to $1120 \times 1120$ pixels, thereby improving its text recognition and GUI comprehension abilities.


To further boost the visual understanding capability, we proposed the \textbf{CogVLM2} (CogVLM2-LLaMA3-8B)
model based on LLaMA3-8B and the \textbf{GLM-4V-9B}
model based on GLM-4-9B. Both CogVLM2 and GLM-4V share nearly the same enhanced training recipe, while the only difference is that CogVLM2 adopts visual expert to preserve language capability while GLM-4V uses text-image co-training. CogVLM2 has surpassed many latest closed-source models in natural image understanding and OCR capabilities, achieving top scores on various benchmarks including MMBench~\cite{MMBench}, VCR~\cite{zhang2024vcr}, MM-Vet~\cite{yu2023mm}, TextVQA~\cite{singh2019towards}, DocVQA~\cite{mathew2021docvqa}, and ChartQA~\cite{masry2022chartqa}. GLM-4V-9B complemented CogVLM2's performance, excelling in OCRbench with the highest score among all models. It also showed strong results in MMStar~\cite{chen2024we}, AI2D~\cite{Ai2D} and MMMU~\cite{yue2023mmmu}, often outperforming other open-source models. These results demonstrate the versatility and effectiveness of both models across a wide range of visual understanding tasks.

To extend CogVLM family's modality to video, recently we proposed \textbf{CogVLM2-Video}
. We propose an automated temporal grounding data construction method based on visual models, and introduce multi-frame video images and timestamps as encoder inputs. CogVLM2-Video not only achieves state-of-the-art performance on public video understanding benchmarks but also excels in video captioning and temporal grounding, providing a powerful tool for subsequent tasks such as video generation and video summarization.

In summary, the research and advancements of the CogVLM model family focus on, but are not limited to, the following aspects:
\begin{itemize}
    \item \textbf{Enhanced vision-language fusion}. Vision-language fusion is a critical topic in VLM training. In VisualGLM~(2023.5), we utilize Qformer as the only trainable parameters to align the image and language spacer. To achieve deeper vision-language alignment without compromising language performance, we designed the Visual Expert architecture and employed it in CogVLM and CogVLM2~(2023.10). Aiming to develop natively multimodal models and promote further multimodal integration, we are actively exploring mixed vision-language training of text and image data such as in GLM-4V-9B~(2024.6).
    
    \item \textbf{Higher input resolution with efficient architecture}. While there is a widespread demand for high-resolution image understanding such as fine-grained image recognition and document understanding, it often results in prohibitive memory and computational costs.
    To overcome this, we proposed efficient high-resolution cross-module in the design of CogAgent~(2023.12), increasing the input resolution of general-domain VLM to $1120\times 1120$ for the first time. We further investigated post-downsample, i.e. downsample the output feature of image encoder, in the training of CogVLM2, GLM-4V and CogVLM2-Video~(2024.5), and found that a $2\times2$ post-downsample with convolutional operator results in almost no noticeable performance degradation, allowing the image sequences to be further shortened.
    
    \item \textbf{Broader modalities and applications}. Since our research on CogVLM, we have progressively being aware of the powerful visual comprehension capabilities of VLM. Therefore, we are dedicated to extending its applications and modalities further, such as visual grounding in CogVLM-grounding~(2023.10), GUI agent in CogAgent~(2023.12) and video understanding~(2024.7). 
\end{itemize}

It is noteworthy that we have open-sourced all the aforementioned VLM models to the research community. 
We hope that our efforts can provide a foundation and offer insights to ongoing research and development in the field of VLM.



%% file: 2_models.tex
\section{The CogVLM2 Family}
The CogVLM family of models comprises four key components: a Vision Transformer (ViT~\cite{alexey2020image}) encoder, an adapter, a language model, and an optional visual expert module. Table~\ref{table:model_compare} provides a comparison of different model architectures within the CogVLM series. Figure~\ref{fig:cogvlm2_video} shows the overall architecture of all CogVLM2 models including CogVLM2, CogVLM2-Video and GLM-4V. 

The ViT encoder transforms discrete raw image inputs into continuous image features rich in semantic content. We employ the state-of-the-art EVA-CLIP~\cite{sun2023eva} model as our image encoder, which demonstrates robust performance across various language model sizes and architectural designs.

The adapter serves as a bridge between visual and linguistic features. While existing approaches like BLIP-2 \cite{li2023blip} and Qwen-VL \cite{bai2023qwen} utilize Q-former for feature alignment, and models such as LLaVA \cite{liu2024visual} and PaLI \cite{chen2022pali} implement a linear layer for this purpose, both methods have limitations. Q-former significantly reduces image sequence length but introduces lossy transformation, sacrificing image details and spatial information. Conversely, LLaVA's linear mapping, though simple and effective, suffers from computational inefficiency due to extended image sequences and the limited expressive capacity of a single linear layer.

To address these challenges, our adapter incorporates a 2$\times$2 convolutional layer followed by a SwiGLU~\cite{shazeer2020glu} module. This design reduces the sequence length output by the ViT to one-quarter of its original size through the convolutional layer. Subsequently, the SwiGLU module aligns these features with linguistic representations, achieving a near-lossless transformation that preserves critical image information while enhancing computational efficiency.


\paragraph{CogVLM2} Similar to the first generation of CogVLM, CogVLM2 features the architecture of visual expert in both the attention and FFN module. This architectural innovation facilitates a deep fusion of visual and linguistic features while preserving the model's inherent language capabilities. For a comprehensive analysis of this implementation, readers are referred to the CogVLM study \cite{wang2023cogvlm}. Different from the first generation model, CogVLM2 further adopts 2$\times$2 downsampling module to increase input resolution while preserving efficiency, and using LLaMA3-8B as the LLM backbone. Besides, we continuously enhance the pre-training as well as post-training data from both diversity and quality aspects, which is detailed in Sec~\ref{subsec:pretrain-data} and Sec~\ref{subsec:posttrain-data}.


\input{tables/archi}




\paragraph{CogVLM2-Video} Currently, the mainstream approach in video understanding involves using image encoders to extract frames from videos, encoding them, and then designing encoding compression modules (e.g., temporal pooling~\cite{maaz2023videovhatgpt,xu2024pllava} or Q-Former modules~\cite{zhang2023video,li2023mvbench}) to compress the video encoding information before inputting it into a large language model (LLM) for joint understanding with textual inputs. Although this method effectively compresses video information, it causes the model to lose temporal awareness, preventing it from accurately associating video frames with precise timestamps. Consequently, the model lacks the capability for temporal localization, timestamp detection, and summarizing key moments. Additionally, video understanding models trained with existing temporal grounding annotated data are limited by the data's scope and the fixed format of question-answering~\cite{wang2022negative,luo2023towards,ren2023timechat,lin2023univtg}, resulting in a lack of open-domain question-answering and processing capabilities. To address these issues, we propose CogVLM2-Video, an extended video model based on the CogVLM2 image understanding model. This model not only achieves state-of-the-art performance in open-domain question-answering but also perceives timestamp information within videos, enabling temporal localization and related question-answering. Specifically, we extract frames from the input video segments and annotate them with timestamp information, allowing the subsequent language model to accurately know the exact time each frame corresponds to in the original video. 

\paragraph{GLM-4V} With the similar training recipe and model architecture as CogVLM2, we further propose GLM-4V, a 13-billion-parameter bilingual visual language model to explore the image understanding capabilities in both English and Chinese. GLM-4V-9B is pre-trained based on GLM-4-9B~\cite{glm2024chatglm}, a recently released open-sourced pre-trained bilingual language model by Zhipu AI. GLM-4V's architecture is similar to CogVLM2, and accommodates input images with a resolution of $1120\times1120$ pixels. The input images are first patchified and processed by a 4B-parameter ViT (EVA-E), downsampled by $2\times2$, then concatenated with language embeddings and fed into the language decoder. We chose a large-scale ViT because we found it critical to the performance of Chinese character recognition. To reduce deployment and computation costs, we preserved the model's language knowledge by image-language co-training instead of utilizing vision experts. However, we observed that the loss of language and image understanding exhibited a competitive relationship, inevitably leading to a slight degradation in the performance of language tasks. Additionally, we present GLM-4V-Plus models, pre-trained for both image and video understanding tasks using the same training recipe. GLM-4V-Plus achieved state-of-the-art performance on a series of image and video understanding benchmarks. Please refer to Table \ref{eval} and Table \ref{vcgbench} for more details.

\input{figures/cogvlm2_video}

%% file: tables/archi.tex
\begin{table}[htbp]
    \caption{Comparison of different model architectures. CA denotes cross attention.}
\begin{center}
\resizebox{\textwidth}{!}{%
        \begin{tabular}{ccccccc}
            \toprule
             \textbf{Model} & \textbf{Parameters} & \textbf{Image Encoder} & \textbf{LLM Backbone} & \textbf{Adapter} & \textbf{Visual Expert}  \\
            \midrule
            VisualGLM & 7B & EVA-CLIP-G & ChatGLM-6B & Q-former & \usym{2717} \\
            CogVLM & 17B & EVA-CLIP-E & Vicuna-1.5-7B & SwiGLU & \usym{2713}\\
            CogAgent & 18B & EVA-CLIP-L/E& Vicuna-1.5-7B & CA + SwiGLU & \usym{2713} \\
            CogVLM2 & 19B & EVA-CLIP-E& LLaMA3-8B & Conv + SwiGLU & \usym{2713} \\  
            GLM-4V-9B & 13B & EVA-CLIP-E& GLM4-9B & Conv + SwiGLU & \usym{2717} \\ 
            CogVLM2-Video & 12B & EVA-CLIP-E & LLaMA3-8B & Conv + SwiGLU & \usym{2717}\\
            \bottomrule
        \end{tabular}
    }
    \end{center}
    \label{table:model_compare}
\end{table}

%% file: figures/cogvlm2_video.tex
\begin{figure}[t]
\begin{center}
\includegraphics[width=0.99\linewidth]{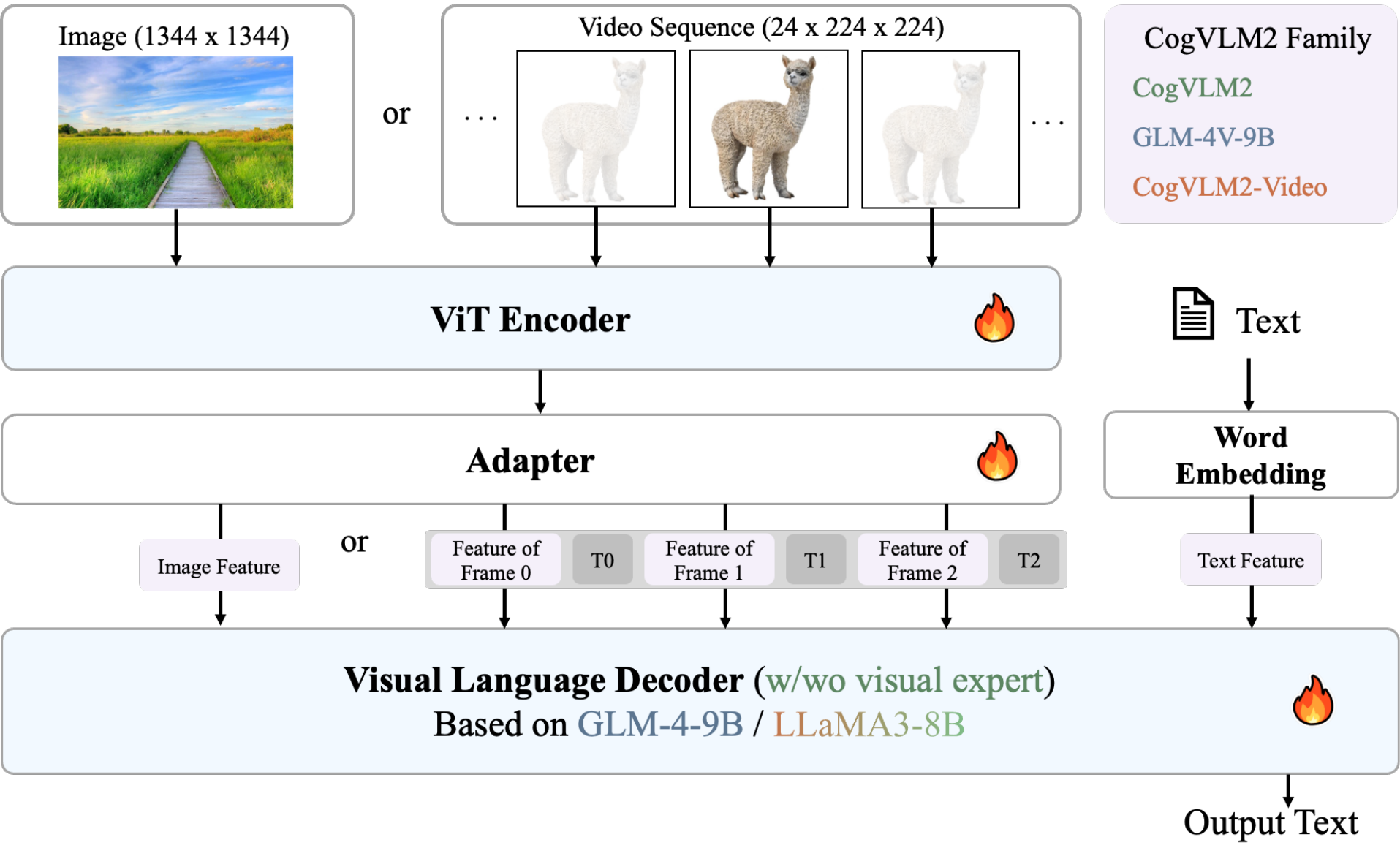}
\end{center}
\caption{The architecture of the CogVLM Family. Taking a high resolution image or the extracted frames from a given video, CogVLM models embed visual information with a pre-trained ViT Encoder and an Adapter. The embedded visual features are sent to a Visual Language Decoder. CogVLM2-Video is capable of answering image-related and video-related queries.}
\label{fig:cogvlm2_video}
\end{figure}

%% file: 3_training.tex
\section{Pre-training}


Pre-training is a foundational phase in the training of VLMs, designed to enhance the model's fundamental ability to understand complex multimodal data. This section provides a comprehensive overview of the methodologies and datasets employed in the pre-training of the CogVLM family. We first discuss our techniques for data processing and generation, including iterative refinement and synthetic data generation, then elaborate on the progressive construction of our pre-training datasets. Furthermore, we explore various pre-training settings that balance the integration of visual and language modalities to achieve optimal model performance across a diverse array of tasks.

\subsection{Pre-training Data}\label{subsec:pretrain-data}

The aim of visual language pre-training is to endow models with the capability to comprehend visual input and align with language space based on large-scale image-text pairs. While there are several open-source large-scale image-text pair datasets, such as LAION~\cite{schuhmann2022laion} and DataComp~\cite{gadre2024datacomp}, they generally contain significant noise and obtaining high-quality image-text pairs is challenging. Additionally, these datasets focus on coarse-grained natural language descriptions of real images, resulting in limited distribution. To address this, we employs two main techniques to obtain and process the pre-training dataset:

\textbf{Iterative Refinement.} While large-scale image-text datasets provide with massive visual language knowledge, they are often noisy or weakly related. Therefore, we use iterative refinement to enhance the data quality. To begin with, the initial model is trained on publicly available datasets, and then used to re-annotate a new batch of data. The annotations generated by the model undergo meticulous manual correction to ensure their accuracy. The corrected data is subsequently used to iteratively refine and enhance future versions of the model. This iterative process fosters continuous improvement in the quality of the training data and, consequently, the model’s performance.

\textbf{Synthetic Data Generation.} The large-scale image-text datasets often focus on coarse-grained natural language descriptions of real images, resulting in limited distribution. For example, they commonly lack data for Chinese text recognition and GUI image understanding. To endow models with a more diverse range of fundamental visual capabilities, we create part of the datasets by synthesizing data according to specific rules or utilizing advanced tools to generate high-quality image-text pairs. 

Utilizing these two techniques, the construction of pre-training data for CogVLM family is progressive and incremental. Here we presents the datasets and their usage in chronological order: 

\textbf{LAION-2B and COYO-700M~\cite{kakaobrain2022coyo-700m}} are two extensive, publicly available datasets comprising numerous images paired with corresponding captions. These datasets form the foundational base for the pre-training stages of all models in CogVLM family, offering a diverse collection of image-text pairs essential for effective model training.

\textbf{LAION-40M-grounding} is an in-house grounding dataset developed using LAION-400M~\cite{schuhmann2021laion} and GLIPv2~\cite{zhang2022glipv2}. This specialized dataset is designed to enhance the model's grounding capabilities, making it particularly suitable for use in models such as CogVLM-grounding and CogAgent, which require precise and accurate grounding annotations.

The \textbf{Digital World Grounding Dataset} consists of 7 million English and 5 million Chinese entries. This dataset is created by crawling web pages with a web browser, capturing screenshots along with all visible DOM elements and their corresponding rendered boxes using Playwright~\footnote{https://playwright.dev}. This comprehensive approach allows for the creation of REC (Referring Expression Comprehension) and REG (Referring Expression Generation) question-answer pairs, significantly enhancing the model’s ability to understand and generate natural language descriptions for visual elements.

The \textbf{Synthetic OCR Dataset} is another vital component of the pre-training data. This dataset includes 120 million English and 150 million Chinese entries, focusing on four specific OCR scenarios: (1) fully generated OCR images with source text printed on the images using Python; (2) real-world images with extracted text obtained using PaddleOCR~\cite{li2022pp}; (3) academic papers with extracted LaTeX code by Nougat~\cite{blecher2023nougat}; and (4) HTML or LaTeX code of tables and formulae rendered to images using various tools. This extensive dataset is utilized in models such as CogAgent, CogVLM2, and GLM-4V to enhance their OCR capabilities.

Finally, \textbf{CLAY-1B} is an in-house recaption dataset built upon LAION-2B and COYO-700M. This dataset is developed with the aid of a fine-tuned CogVLM model specifically designed to generate long, detailed captions for images. The Chinese captions in this dataset are translated by a fine-tuned ChatGLM. CLAY-1B is used in models like CogVLM2 and GLM-4V to improve their captioning abilities.

\subsection{Pre-training Settings}

With a pre-trained language model as a starting point, the primary objective during visual language pre-training is to incorporate the image modality into the model while minimizing potential decline in its language abilities. To achieve this, three main visual-language training methodologies are explored:

The first approach involves progressively enabling more trainable parameters as pre-training stages advance. For instance, in the initial stage of CogAgent, only the cross-attention layers are trained. As training progresses, additional parameters, such as those of the Vision Transformer (ViT) or other vision experts, are gradually made trainable. This staged approach ensures that the model can seamlessly integrate the image modality without compromising its pre-existing language capabilities.

The second approach involves training all parameters simultaneously but utilizing both language pre-training data and visual-language pre-training data. This method, particularly adopted in GLM-4V training, ensures that the model is exposed to a balanced mix of data types, thereby maintaining its language abilities while effectively incorporating visual information.

The third approach entails gradually increasing the input image resolution as training progresses. By starting with lower resolution images and progressively enhancing the resolution, the model can adapt to handling higher-quality visual information over time. This gradual increase in resolution allows the model to capture and comprehend the finer details in images, thereby enhancing its overall visual comprehension capabilities.

These pre-training settings are meticulously designed to optimize the integration of visual and language modalities, ensuring that the resulting models can perform effectively across a broad spectrum of tasks involving both text and images.

\section{Post-training}

\subsection{Post-training Data}\label{subsec:posttrain-data}

\begin{table}[!h]
\caption{VQA datasets used in image understanding models. The "Type" column signifies the format of the answers provided. "0" corresponds to concise responses, such as multiple-choice, Y/N, etc. "1" denotes comprehensive answers that incorporate a chain of thought processes.}

\centering
\renewcommand{\arraystretch}{1.3}
\resizebox{\textwidth}{!}{%
\setlength{\tabcolsep}{12pt}
\begin{tabular}{cccccc}
\toprule
\textbf{Categories}            & \textbf{Datasets}                 & \textbf{Type}& \textbf{CogVLM} & \textbf{CogVLM2} & \textbf{GLM4V-9B} \\ \hline
\multirow{6}{*}{General QA}      & OKVQA~\cite{okvqa}              & 0                    & \checkmark  & \checkmark    & \checkmark   \\
                                 & STVQA~\cite{STVQA}              & 0                    &           & \checkmark    & \checkmark   \\
                                 & VGQA~\cite{VGQA}                & 0                    &           & \checkmark    & \checkmark   \\
                                 & VQAV2~\cite{vqav2}              & 0                    & \checkmark  & \checkmark    & \checkmark   \\
                                 & A-OKVQA~\cite{A-OKVQA}          & 0                    &           &             & \checkmark   \\
                                 & TQA~\cite{TQA}                  & 0                    &           &             & \checkmark   \\
                                 \cline{1-6}                   
 \multirow{5}{*}{OCR}            & IAM~\cite{IAM}                  & 1                    &           &             & \checkmark   \\
                                 & DocVQA~\cite{DocVQA}            & 0                    &           & \checkmark    & \checkmark   \\
                                 & OCRVQA~\cite{ocrvqa}            & 0                    & \checkmark  & \checkmark    & \checkmark   \\
                                 & TextVQA~\cite{textvqa}          & 0                    & \checkmark  & \checkmark    & \checkmark   \\
                                 & Rendered\_text~\footnote{\url{https://huggingface.co/datasets/wendlerc/RenderedText}}     
                                 & 0                    &           &             & \checkmark   \\
                                 \cline{1-6}                   
\multirow{7}{*}{Math \& Science} & GeoMetry3K~\cite{InterGPS}    & 0                    &           & \checkmark    & \checkmark  \\
                                 & Geo170K~\cite{Geo170K}          & 1                    &           & \checkmark    & \checkmark   \\
                                 & GeoQA~\cite{GeoQA}              & 0                    &           & \checkmark    & \checkmark   \\
                                 & Geomverse~\cite{Geomverse}      & 1                    &           &             & \checkmark   \\
                                 & Raven~\cite{Raven}              & 1                    &           &             & \checkmark   \\
                                 & InterGPS~\cite{InterGPS}        & 0                    &           &             & \checkmark   \\
                                 & Ai2D~\cite{Ai2D}                & 0                    &           &             & \checkmark   \\ 
                                 & ScienceQA~\cite{scienceqa}      & 1                    & \checkmark  & \checkmark    & \checkmark   \\
                                 \cline{1-6}                   
\multirow{8}{*}{Chart Analysis}  & ChartQA~\cite{ChartQA}          & 0                    &           & \checkmark    & \checkmark   \\
                                 & FigureVQA~\cite{FigureQA}       & 0                    &           & \checkmark    & \checkmark  \\
                                 & InfoVQA~\cite{InfographicVQA}   & 0                    &           & \checkmark    & \checkmark   \\
                                 & DVQA~\cite{DVQA}                & 0                    &           & \checkmark    & \checkmark   \\
                                 & ArxivQA~\cite{ArxivQA}          & 1                    &           & \checkmark    & \checkmark   \\
                                 & TabMWP~\cite{TabMWP}            & 1                    &           &             & \checkmark   \\
                                 & VQARAD~\cite{VQARAD}            & 0                    &           &             & \checkmark   \\ 
                                 \cline{1-6}               
\multirow{6}{*}{Other}           & VSR~\cite{VSR}                  & 0                    &           &             & \checkmark   \\
                                 & TDIUC~\cite{TDIUC}              & 0                    &           & \checkmark    & \checkmark   \\
                                 & TallyQA~\cite{TallyQA}          & 0                    &           & \checkmark    & \checkmark   \\
                                 & IconQA~\cite{IconQA}            & 0                    &           &             & \checkmark   \\
                                 & VisText~\cite{VisText}          & 0                    &           &             & \checkmark   \\
                                 & Diagram\_image\_to\_text~\footnote{\url{https://huggingface.co/datasets/Kamizuru00/diagram_image_to_text}}        
                                 & 1                    &           &             & \checkmark   \\
\bottomrule
\label{vqa-datasets}
\end{tabular} }
\end{table}


\paragraph{Image Post-Training Datasets.} Our image post-training datasets is composed of a collection of open-sourced visual question-answering (VQA) datasets and annotateded  alignment data. 

The whole collection of VQA datasets are listed in \Cref{vqa-datasets}. 
Experimental results show that incorporating more and broader VQA data can continuously and effectively enhance the model's performance. Consequently, in comparison to CogVLM~\cite{wang2023cogvlm}, we added more VQA datasets in CogVLM2 and GLM-4V. During our experiments, we observed that add VQA data with concise responses (such as VQAv2~\cite{vqav2}) could detract from the model's conversational performance. 
To deal with it, we prefixed concise answers with ``Short Answer'' to distinguish VQA-type concise responses from dialogue-type responses, thereby reducing interference between the two.

Moreover, we meticulously annotated approximately 300K alignment corpora. Based on the characteristics of the images and instructions, these corpora are categorized into different categories for proportional control during training. Additionally, we annotated 50K preference alignment corpora to steer the model towards generating outputs that align with human preferences.

\begin{table}[!h]
\caption{Datasets used in video understanding models.}
\centering
\renewcommand{\arraystretch}{1.5}
\resizebox{\textwidth}{!}{%
\setlength{\tabcolsep}{30pt}
\begin{tabular}{cc}
\toprule
\textbf{Categories}                 & \textbf{Datasets}                          \\  \hline
Video Caption      & In-house Detailed Caption Dataset \\
Video QA           & \makecell[c]{VideoChat~\cite{videochat}, VideoChatGPT~\cite{maaz2023videovhatgpt}, NExT-QA~\cite{nextqa}\\CLEVRER~\cite{clevrer}, Kinetics-710~\cite{uniformerv2}, SthSthV2~\cite{sth}\\Ego4D~\cite{ego4d}, TGIF-QA~\cite{tgif_qa}, WebVidQA~\cite{just_ask}\\In-house VideoQA Dataset}                            \\
Temporal Grounding & TQA Dataset \\
\bottomrule
\label{tab: video-datases}
\end{tabular}
}
\end{table}

\input{figures/tqa_data}
\label{para:tqa data}
\paragraph{Video TQA Dataset.} The training of video understanding models using existing temporal grounding annotation data is limited by the scope of the data and the fixed format of question and answer pairs~\cite{wang2022negative,luo2023towards,ren2023timechat,lin2023univtg}, lacking the capability for open-domain question answering and processing. Compared to the plain text data used to train LLMs and the image understanding data used to train VLMs, the annotation cost for high-quality video question answering and temporal grounding data is extremely high. Manual annotation alone cannot meet the demands of large-scale training. To prepare temporal grounding data suitable for large-scale training, we developed a fully automated video question-answering data generation process as shown in Figure~\ref{fig:tqa_data}. We leverage the latest image understanding models to extract frame-level understanding from video data, and then use GPT-4o~\cite{gpt4o} for data filtering and generation. Through this automated data processing workflow and large-scale training, CogVLM2-Video not only excels on public benchmarks but also possesses the temporal question-answering capability that most previous video models lacked. Through this pipeline, we ultimately generated 30k Temporal Grounding Question and Answer (TQA) data points. More details for generating the TQA dataset can be found in Appendix~\ref{appendix:tqa}.

\subsection{Post-training Settings}

\paragraph{Image Supervised Fine-tuning.}

In CogVLM2 and GLM-4V, we employed a two-stage SFT training approach. In the first stage, we utilized all VQA training datasets and the 300K alignment corpora to enhance the model's foundational capabilities, addressing the limitations of pre-training on image captioning tasks. In the second stage, we selected a subset of VQA datasets and the 50K preference alignment data to optimize the model's output style, closely aligning with human preferences. 

In the first stage, the model underwent 3000 iterations with a learning rate of 1e-5 and a global batch size of 2340. Subsequently, in the second stage, we reduced the global batch size to 1150 for 750 steps. We performed the image SFT process by fine-tuning all parameters. To enhance and ensure the stability of the training, we activated the visual encoder’s parameters and adjusted its learning rate to be one-tenth of that used for the remaining training parameters.

\paragraph{Video Supervised Fine-tuning.} Starting from a pre-trained $224\times224$ variant of CogVLM2 image understanding model, CogVLM2-Video takes $24$ frames as input and extract visual information sequentially. We add an additional convolution layer with $2\times2$ kernel at the end of the ViT model to further compress the video features. The training process consists of two stages: instruction tuning and temporal grounding tuning. All the parameters are trainable throughout these two stages. In the instruction tuning stage, in-house detailed caption data and public available question-answering data are utilized to improving the general video understanding capability of the model, with a learning rate of $4e-6$. We mainly used the instruction data provided in VideoChat2, without the simple caption datasets. We also collected an in-house video QA dataset for better temporal understanding. A total of 330k video samples are utilized in the instruction tuning. In the temporal grounding tuning stage, CogVLM2-Video is trained on the TQA Dataset with a learning rate of $1e-6$. The complete training process takes around $8$ hours with a cluster of 8 nodes of NVIDIA A100 machines. For different application senarios, we release two models: "cogvlm2-video-llama3-base" and "cogvlm2-video-llama3-chat". "cogvlm2-video-llama3-base" is the model trained with datasets in stage one, excelling the existing video understanding benchmarks; "cogvlm2-video-llama3-chat" is the model further fine-tuned on the TQA Dataset, which has the temporal grounding capability. Please refer to Table~\ref{tab: video-datases} for detailed video instruction corpus.

%% file: figures/tqa_data.tex
\begin{figure}[t]
\begin{center}
\includegraphics[width=\linewidth]{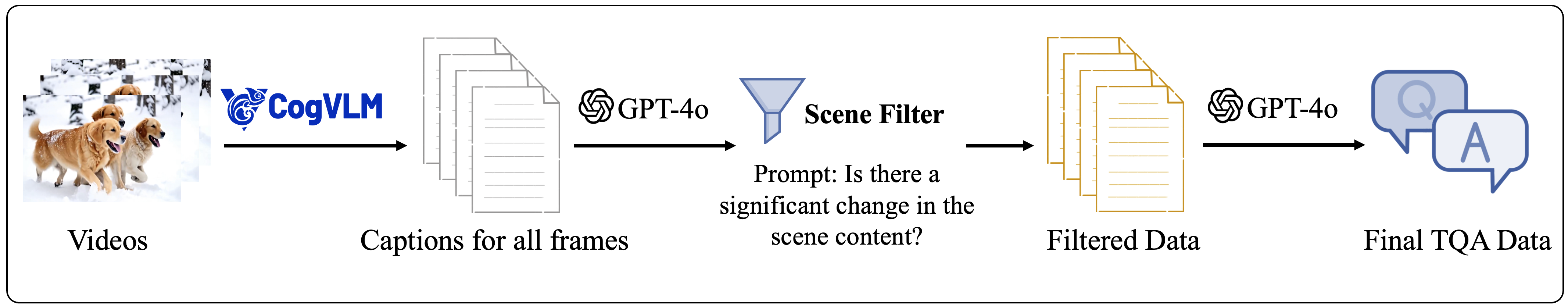}
\end{center}
\caption{The pipeline for temporal grounding data generation. In this pipeline, we first extract frames from the video and then use CogVLM2 for image captioning. Next, we use GPT-4o to evaluate the series of image captions for each video, identifying videos with significant scene content changes and filtering out those with minor scene content changes. Finally, using a few-shot approach, GPT-4o generates time-related question-answer pairs based on the image captions. More details can be found in Appendix~\ref{appendix:tqa}.}
\label{fig:tqa_data}
\end{figure}

%% file: 4_evaluation.tex
\section{Evaluation}
Here we evaluate our CogVLM2 family, including CogVLM2, CogVLM2-Video, GLM-4V-9B. Following the training recipe of GLM-4V-9B and CogVLM2-Video, we further pre-trained GLM-4V-Plus, powerful internal visual language models capable of image and video understanding which are available at Zhipu MaaS platform\footnote{\url{https://bigmodel.cn/}}. Our comprehensive evaluation of the CogVLM2 family spans a wide range of image and video understanding tasks, demonstrating its capabilities across diverse visual domains. 

\input{figures/evaluation}

\subsection{Evaluation of Image Tasks}


We evaluate our models on multiple image tasks alongside widely used large visual language models, including both proprietary and open-source variants, as detailed in Table \ref{eval}. To comprehensively assess the performance of our models, we select the following tasks: (1) OCR comprehension: TextVQA~\cite{singh2019towards}, DocVQA~\cite{mathew2021docvqadatasetvqadocument}, OCRbench~\cite{liu2024hiddenmysteryocrlarge}, VCR~\cite{zhang2024vcr}); (2) Chart and diagram understanding: ChartQA~\cite{masry2022chartqabenchmarkquestionanswering}, AI2D~\cite{kembhavi2016diagram}; (3) Subject-specific question answering: MMMU~\cite{yue2023mmmu}; (4) General question answering: MMVet~\cite{yu2024mm}, MMBench~\cite{MMBench}, MMStar~\cite{chen2024we} and MME~\cite{yin2023survey}. 
Compared to open-source models of similar parameter scales, CogVLM2 and GLM-4V-9B achieve state-of-the-art performance on most tasks, and even surpass models of much larger scale such as Mini-Gemini 34B, LLaVA-NeXT-110B and proprietary models such as QwenVL-Plus, Claude3-Opus, Gemini 1.5 Pro, GPT-4v-20231106 on multiple benchmarks. 

\begin{table}[htbp]
\caption{\small Image understanding performance comparison on popular benchmarks. The best results are \textbf{bolded}, and the second highest results are \underline{underlined}. 
}

\label{eval}
\centering
\resizebox{\textwidth}{!}{%
\setlength{\tabcolsep}{1mm}{
\begin{tabular}{lccccccccccc}
\toprule
Model & MMBench & MMStar & MMVet & MMMU & MME & ChartQA & AI2D & TextVQA & DocVQA & VCR & OCRbench \\
\hline
\multicolumn{11}{c}{\textit{Proprietary models}} \\
\hline
Step-1V & 80.7 & 50.0 & 63.3 & 49.9 & - & 57.7 & 79.2 & 71.6 & - & - & 625 \\
Qwen-VL-Max\cite{Qwen-VL} & 77.6 & 49.5 & 61.8 & 52.0 & \underline{2281} & 79.8 & 75.7 & 79.5 & \textbf{93.1} & 76.8 & 684 \\
Claude3.5-Sonnet & 79.7 & 62.2 & 66 & 65.9 & 1920 & - & 80.2 & - & - & 63.9 & 788 \\
Gemini 1.5 Pro\cite{reid2024gemini} & 73.9 & 59.1 & 64.0 & 60.6 & 2110 & 81.3 & 79.1 & 78.7 & 86.5 & 62.7 & 754 \\
GPT-4v-20231106\cite{GPT4} & 77.0 & 49.7 & 56.8 & 53.8 & 1771 & - & 75.9 & - & - & 52.0 & 516 \\
GPT-4o-20240513 & \underline{83.4} & \textbf{63.9} & \underline{69.1} & \textbf{69.2} & \textbf{2310} & \textbf{85.7} & \textbf{84.6} & - & 92.8 & \textbf{91.6} & 736 \\
\hline
\multicolumn{11}{c}{\textit{Open-source models}} \\
\hline
CogVLM1.1\cite{wang2023cogvlm} & 65.8 & 39.9 & 52.0 & 37.3 & 1737 & 68.3 & 63.3 & 69.7 & - & - & 590 \\
Cambrian-34B\cite{tong2024cambrian1} & 80.4 & 54.2 & 53.2 & 50.4 & 2050 & 73.7 & 79.5 & 72.1 & 75.5 &  79.7 & 591 \\
LLaVA-V1.5-13B\cite{liu2024improvedbaselinesvisualinstruction} & 69.2 & 34.3 & 36.3 & 37.0 & 1781 & 18.2 & 61.1 & 48.9 & - & - & 337 \\
Mini-Gemini\cite{li2024minigemini} & 80.6 & - & 59.3 & 48.0 & - & - & - & 74.1 & - & - & - \\
MiniCPM-Llama3-V2.5\cite{yao2024minicpm} & 77.6 & 51.8 & 52.8 & 45.8 & 2025 & - & 78.4 & 76.6 & 84.8 & 31.8 & 725 \\
InternVL2-26B\cite{chen2023internvl} & \underline{83.4} & 61.0 & 60.0 & 50.7 & 2259 & \underline{84.9} & \underline{84.5} & 82.5 & \underline{92.9} & 74.5 & \underline{825} \\
LLaVA-Next-Yi-34B\cite{liu2024llavanext} & 81.1 & 51.6 & 50.7 & 48.8 & 2006 & 67.6 & 78.9 & 69.3 & - & - & 574 \\
\hline
\multicolumn{11}{c}{\textit{Ours}} \\
\hline
CogVLM2-LLaMA3 & 80.5 & 50.5 & 60.4 & 44.3 & 1870 & 81.0 & 73.4 & \underline{84.2} & 92.3 & \underline{83.3} & 756 \\
GLM-4V-9B & 81.1 & 58.7 & 58.0 & 47.2 & 2164 & 71.1 & 81.1 & 83.0 & 81.0 & 43.7 & 786 \\
GLM-4V-Plus & \textbf{84.6} & \underline{62.9} & \textbf{71.1} & 53.0 & 2275 & 83.1 & 83.9 & \textbf{86.6} & 92.3 & - & \textbf{833} \\
\bottomrule
\end{tabular}}}
\end{table}

\subsection{Evaluation of Video Tasks}
CogVLM2-Video achieves state-of-the-art performance on multiple video question-answering tasks. Table \ref{vcgbench} shows the performance of CogVLM2-Video on the MVBench~\cite{li2023mvbench}, VideoChatGPT-Bench~\cite{Maaz2023VideoChatGPT} and LVBench~\cite{wang2024lvbench}. Where MV-* refers to the MVBench, VCG-* refers to the VideoChatGPT-Bench. More details on MVBench can be found in Appendix~\ref{ap:mvbench}.

\begin{table}[htbp]
\scriptsize
\caption{\small Video understanding performance comparison on MVBench~\cite{li2023mvbench}, VideoChatGPT-Bench~\cite{Maaz2023VideoChatGPT} and LVBench~\cite{wang2024lvbench}. The best results are \textbf{bolded}, and the second highest results are \underline{underlined}.}
\label{vcgbench}
\centering
\resizebox{\textwidth}{!}{%
\setlength{\tabcolsep}{1mm}{
\begin{tabular}{lccccccccccc}
\toprule[0.85pt]
Model                                                   & MV-AVG            &  MV-Action            &  MV-Object            &   MV-Cognition        & VCG-AVG       & VCG-CI        & VCG-DO        & VCG-CU        & VCG-TU        & VCG-CO        & LVBench         \\\hline
IG-VLM GPT4V~\cite{kim2024imagegridworthvideo}          & 43.7              &   62.2                &  35.7                 &   33.7                & 3.17          & 3.40          & 2.80          & 3.61          & 2.89          & 3.13          & -       \\
ST-LLM~\cite{liu2023one}                                & 54.9              &   61.2                &  64.2                 &   44.8                & 3.15                 & 3.23          & 3.05          & 3.74          & 2.93          & 2.81          & -          \\
ShareGPT4Video~\cite{chen2024sharegpt4video}            & 51.2              &   52.6                &  56.5                 &   39.7                & -                    & -           & -           & -           & -           & -           & -          \\
VideoGPT+~\cite{maaz2024videogptintegratingimagevideo}  & 58.7              &   65.4                &  65.7                 &   44.5                & 3.28          & 3.27          & 3.18          & 3.74          & 2.83          & \underline{3.39} & -          \\
VideoChat2\_HD\_mistral~\cite{li2023mvbench}            & \underline{62.3}  &   63.4                &  53.5                 &   51.0                & 3.10          & 3.40          & 2.91          & 3.72          & 2.65          & 2.84          & -          \\
PLLaVA-34B~\cite{xu2024pllava}                          & 58.1              &   65.7                &  57.5                 &   \underline{54.0}    & 3.32          & \textbf{3.60} & 3.20          & \underline{3.90} & 2.67          & 3.25          & 26.1 \\
GPT-4o-2024-05-13~\cite{gpt4o}                          & 47.8              &   56.7                &  50.5                 &   43.0                & -          & - & -         &- & -          & -        & \underline{34.7} \\
Gemini 1.5 Pro~\cite{reid2024gemini}                    & 52.6              &   58.5                &  57.0                 &   45.8                & -          & - & -         &- & -          & -        & 33.1 \\
\hline
CogVLM2-Video (ours)                                    & \underline{62.3}  &   \underline{70.6}    &  \underline{68.3}     &   49.3                & \underline{3.41} & 3.49          & \underline{3.46} & 3.87          & \textbf{2.98} & 3.23          & 28.1      \\
GLM-4V-Plus (ours)                                      & \textbf{71.2}     &   \textbf{72.0}       &  \textbf{71.0}        &   \textbf{57.3}       & \textbf{3.51} & \underline{3.57}          & \textbf{3.55} & \textbf{3.95}          & \underline{2.92} & \textbf{3.55}         & \textbf{38.3}   \\
\bottomrule[0.85pt] 
\end{tabular}}}
\end{table}

%% file: figures/evaluation.tex
\begin{figure*}[h]
\center
\subfloat[Evaluation results on image tasks]{\begin{centering}
\includegraphics[width=0.5\linewidth]{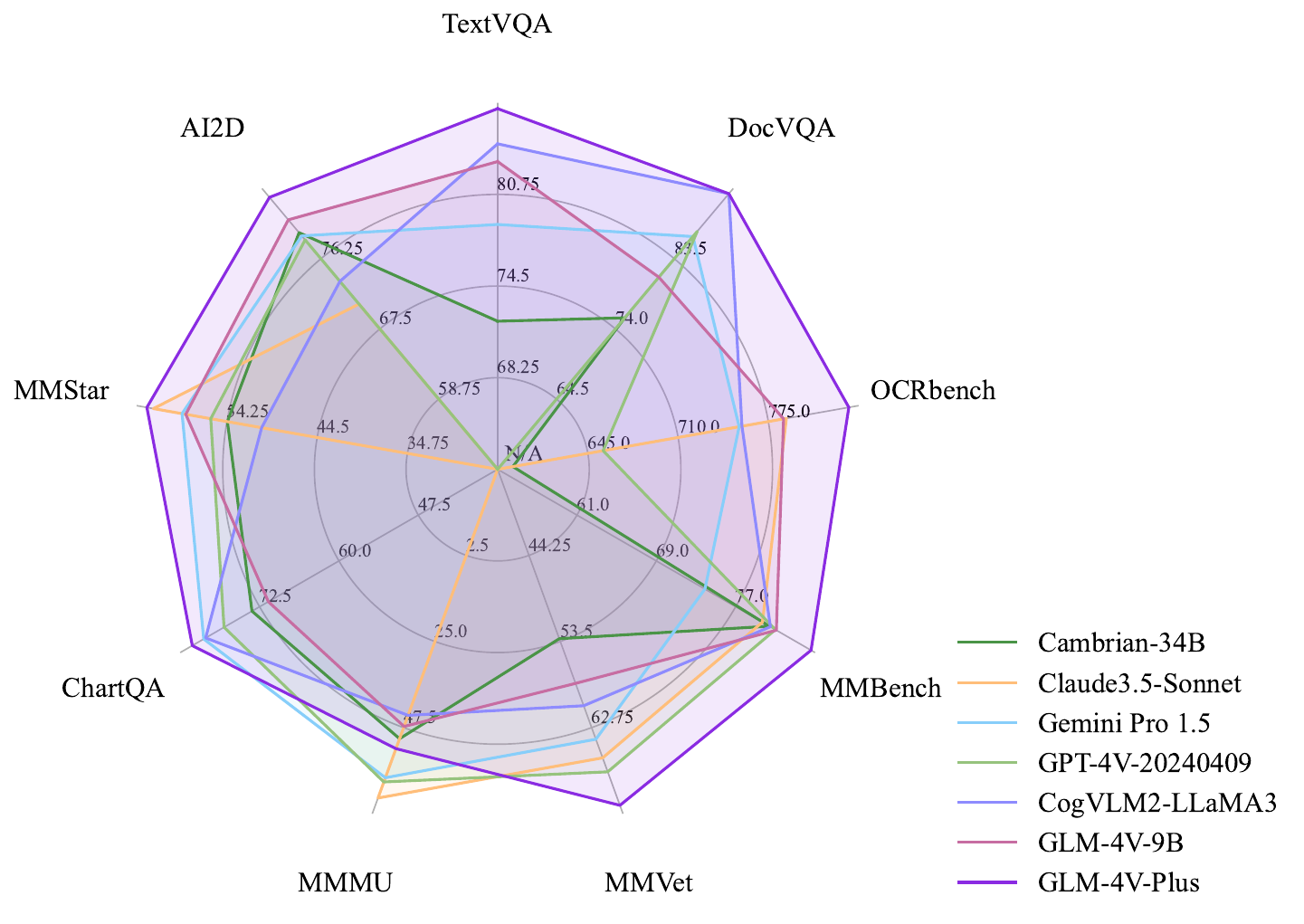}
\end{centering}
}\subfloat[Evaluation results on video tasks.]{\begin{centering}
\includegraphics[width=0.5\linewidth]{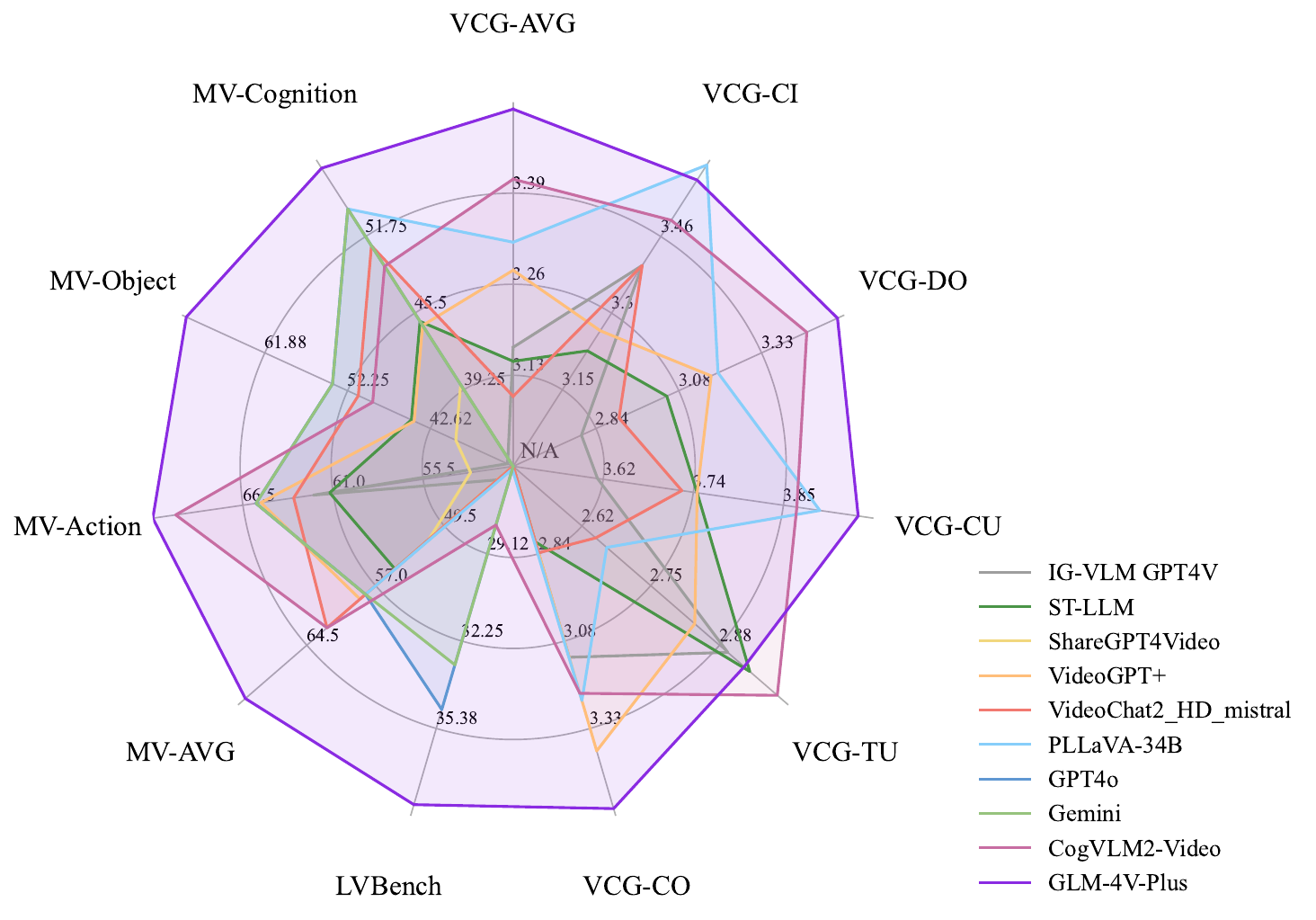}
\end{centering}
}

\caption{Performance visualization of CogVLM2 family with radar charts.}
\label{fig:eval}
\end{figure*}

%% file: 5_discussion.tex
\section{Conclusion}
The CogVLM Family represents a significant advancement in the integration of visual and language modalities, addressing limitations of traditional LLMs that are restricted to text-based inputs. Here we propose and open-source the CogVLM2 family. By introducing models capable of understanding and generating content from both images and videos, the CogVLM series expands the potential applications of LLMs in various domains, including document analysis, GUI comprehension, and temporal video grounding. The architectural innovations, such as the Visual Expert and high-resolution cross-modules, enable a seamless fusion of visual and linguistic features, enhancing the models' performance without compromising their language capabilities. Furthermore, the efficient use of high-resolution inputs and sophisticated data generation techniques ensures that the models are well-equipped to handle complex visual-language tasks. Future research could explore even broader modalities and improved alignment techniques to further enhance the capabilities of VLMs. Overall, the CogVLM Family sets a new benchmark for open-source VLMs, providing powerful tools for both academic research and practical applications.

%% file: appendix/TQA_details.tex
\section{Details for Generating TQA Dataset}
\label{appendix:tqa}

In the generation pipeline, we first extract frames from the video and then use CogVLM2 for image captioning. The prompt for the image caption is shown below:
\begin{promptbox}[Prompt for Image Caption]
\noindent
Give out the detailed description of this image.
\end{promptbox}

Next, we use GPT-4o to evaluate the series of image captions for each video, identifying videos with significant scene content changes and filtering out those with minor scene content changes. The prompt for the scene filter is shown below:

\begin{promptbox}[Prompt for Scene Filter]
\noindent
\begin{verbatim}
We extracted several frames from this video and described each frame 
using an image caption model, stored in the dictionary variable 
`image_captions: Dict[str:str]`. In `image_captions`, the key is the 
second at which the image appears in the video, and the value is a 
detailed description of the image at that moment. Our image captions may 
contain hallucinations and errors. If you find any information that 
seems incorrect, please ignore the erroneous information. 
image_captions={images_caption}
Please determine whether there are significant scene changes in each 
second of the video based on the frame descriptions. If there are 
significant changes, output \"Yes\"; otherwise, output \"No\".For 
example, if the description continuously mentions a pool scene or a 
kitchen scene, then return \"No\". If the description first mentions an 
indoor scene and then a beach scene, then return \"Yes\". Output your 
final answers directly. Do not give out your reason.
\end{verbatim}
\end{promptbox}

Finally, using a few-shot approach, GPT-4o generates time-related question-answer pairs based on the image captions. The prompt for question-answer generation is shown below:

\begin{promptbox}[Prompt for Question-Answer Generation]
\noindent
\begin{verbatim}
We extracted several frames from this video and described each frame 
using an image caption model, stored in the dictionary variable 
`image_captions: Dict[str:str]`. In `image_captions`, the key is the 
second at which the image appears in the video, and the value is a 
detailed description of the image at that moment. You will play two 
roles: a human asking a question related to the biggest change in the 
video and an intelligent chatbot designed for video understanding.Your 
question and answer should be based on the information provided by 
`image_captions`.\n" "image_captions={images_caption}\nPlease generate 
the response in the form of a Python dictionary string with keys 
\"Human\" for question and \"Bot\" for answer. Each corresponding value 
should be the question and answer text respectively. For example, your 
response should look like this: {\"Human\": \"Your question here...\", 
\"Bot\": \"Your answer here...\"}. Do not mention variables like 
`image_captions` in your response. Always generate the question and the 
answer as if you are directly looking at the video.The questions you 
generate should focus on the biggest scene change. Do not generate 
questions that can be easily answered from a single image. Here are some 
examples of what we expect:\nExample 1: {\"Human\": \"At what second 
does the girl appear?\", \"Bot\": \"The girl appears at the 3rd second 
in the video.\"}\nExample 2: {\"Human\": \"When does the video switch 
from the swimming pool to the grass?\", \"Bot\": \"At the 3rd 
second.\"}\nExample 3: {\"Human\": \"When does a significant scene 
change occur in the video? \", \"Bot\": \"At the 10th second, the scene 
changes from the playground to the classroom.\"}\nIf you think it is not 
possible to generate such questions and answers from the given image 
captions, output \"None\" directly. Keep your answers concise and 
accurate and generate only one question and answer pair. 
\end{verbatim}
\end{promptbox}

%% file: appendix/CogVLM2_cases.tex
\section{Qualitative Examples}
\label{appendix:cases}


This section will demonstrate the capabilities of CogVLM2 in various scenarios. Figures \ref{fig:understanding1}, \ref{fig:understanding2}, \ref{fig:understanding3}, \ref{fig:understanding4},\ref{fig:understanding5} Show the perception and understanding ability of CogVLM2, while Figures \ref{fig:reasonging1}, \ref{fig:reasonging2}, \ref{fig:reasonging3}, \ref{fig:reasonging3},\ref{fig:reasonging4}
exhibit Reasoning and Analysis.

\subsection{Perception and Understanding}

\begin{promptbox}[Text Recognition]
\noindent
\textcolor{blue}{Prompt:} \\
\begin{CJK*}{UTF8}{gbsn}
请详细描述此户型图，包括各个房间的名称和大小。
\end{CJK*}
\begin{figure}[H] %
  \centering
\includegraphics[width=\linewidth]{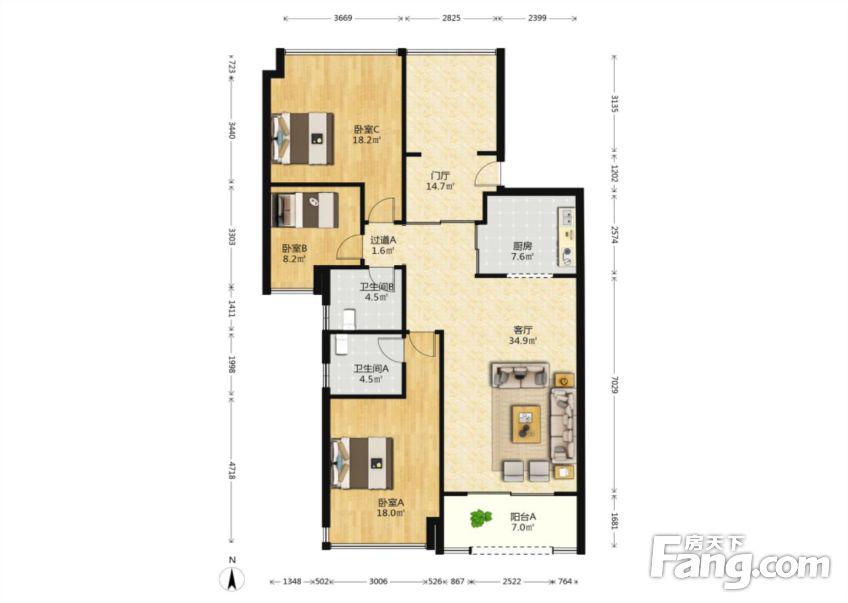}
  \caption{Image Case 1}
  \label{fig:understanding1} 
\end{figure}

\noindent 
\textcolor{blue}{CogVLM2:} \\
\begin{CJK*}{UTF8}{gbsn}
户型图显示了一个住宅的平面布局。从左到右，上到下，我们可以看到如下房间和区域：\\
- 卧室C：位于左侧，面积为18.2平方米。\\
- 门厅：位于中间，面积为14.7平方米。 \\
- 卧室B：位于左侧，面积为8.2平方米。 \\
- 过道A：连接卧室B和门厅，面积为1.6平方米。\\
- 厨房：位于右侧，面积为7.6平方米。\\
- 卫生间B：位于门厅旁边，面积为4.5平方米。\\
- 客厅：位于右侧，面积为34.9平方米。\\
- 卫生间A：位于客厅旁边，面积为4.5平方米。 \\
- 卧室A：位于左侧，面积为18.0平方米。 \\
- 阳台A：位于右侧，面积为7.0平方米。\\
在户型图的左下角还有一个表示方向的罗盘，上面标注了“N”代表北。此外，图片的右下角有一个水印，显示为“房天下Fang.com”。
\end{CJK*}
\end{promptbox}

\begin{promptbox}[Text Recognition]
\noindent
\textcolor{blue}{Prompt:} \\
\begin{CJK*}{UTF8}{gbsn}
请详细描述此图像内容。
\end{CJK*}
\begin{figure}[H] %
  \centering
\includegraphics[width=0.9\linewidth]{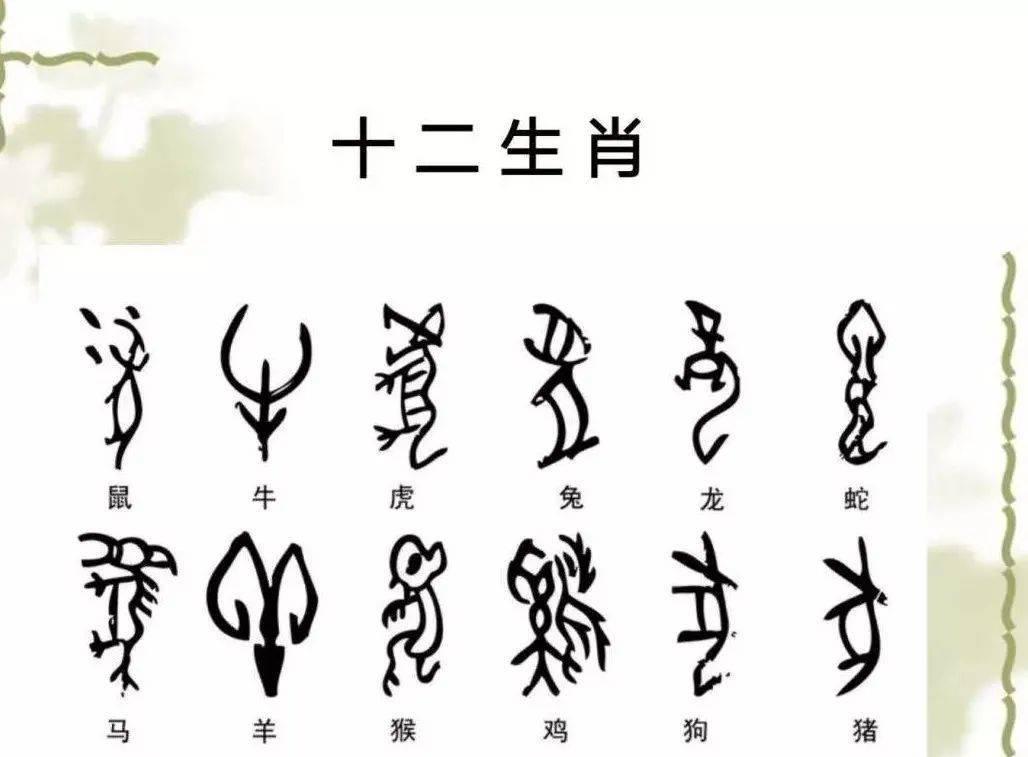}
  \caption{Image Case 2}
  \label{fig:understanding2} 
\end{figure}

\noindent 
\textcolor{blue}{CogVLM2:} \\
\begin{CJK*}{UTF8}{gbsn}
这是一幅展示了十二生肖的图片。图片上方有“十二生肖”四个大字，下方则是具体的十二生肖图案，每个生肖都由一个独特的图形表示。这些图形与常见的生肖符号相似，但又有所不同，可能是对传统生肖的现代艺术诠释。从左到右、从上到下，十二生肖依次为：鼠、牛、虎、兔、龙、蛇、马、羊、猴、鸡、狗、猪。
\end{CJK*}
\end{promptbox}

\begin{promptbox}[Logo Recognition]
\noindent
\textcolor{blue}{Prompt:} \\
Describe the image
\begin{figure}[H] %
  \centering
\includegraphics[width=\linewidth]{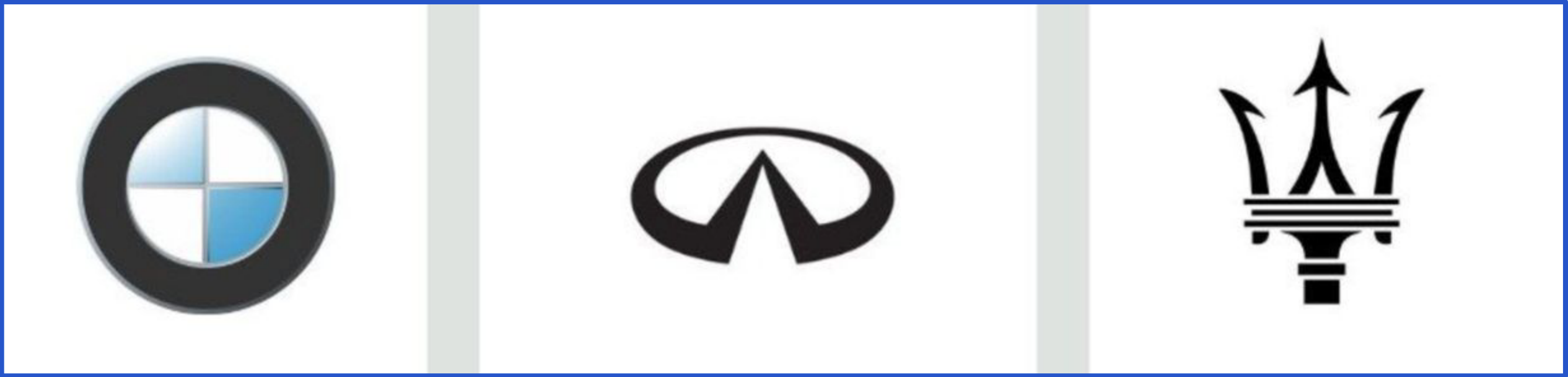}
  \caption{Image Case 3}
  \label{fig:understanding3} 
\end{figure}

\noindent 
\textcolor{blue}{CogVLM2:} \\
The image displays three distinct logos arranged in a row. From left to right, the logos are:
   
   1. The first logo is for BMW. It features a blue and white circular design with a black border. The blue and white colors are arranged in a quadrant pattern, with the blue sections forming a cross in the center. The BMW logo is a recognizable symbol for the German automobile manufacturer.
   
   2. The second logo is for Infiniti, a luxury vehicle division of Nissan. It consists of a stylized letter "Q" enclosed within a circular shape. The letter is split into two parts, with the top half pointing upwards and the bottom half pointing downwards, creating a unique and modern design.
   
   3. The third logo is for Maserati, an Italian luxury car manufacturer. It depicts a trident, which is a symbol of the Roman god Neptune, with the trident's prongs pointing upwards. The trident is set against a black background, and the overall design is sleek and elegant, embodying the luxury and heritage of the Maserati brand.
\end{promptbox}

\begin{promptbox}[Information Extraction]
\noindent
\textcolor{blue}{Prompt:} \\
Please read the text in this image and return the information in the following JSON format (note xxx is placeholder, if the information is not available in the image, put "N/A" instead).\{"Surname": xxx, "Given Name": xxx, "USCIS \#": xxx, "Category": xxx, "Country of Birth": xxx, "Date of Birth": xxx, "SEX": xxx, "Card Expires": xxx, "Resident Since": xxx\}
\begin{figure}[H] %
  \centering
\includegraphics[width=0.8\linewidth]{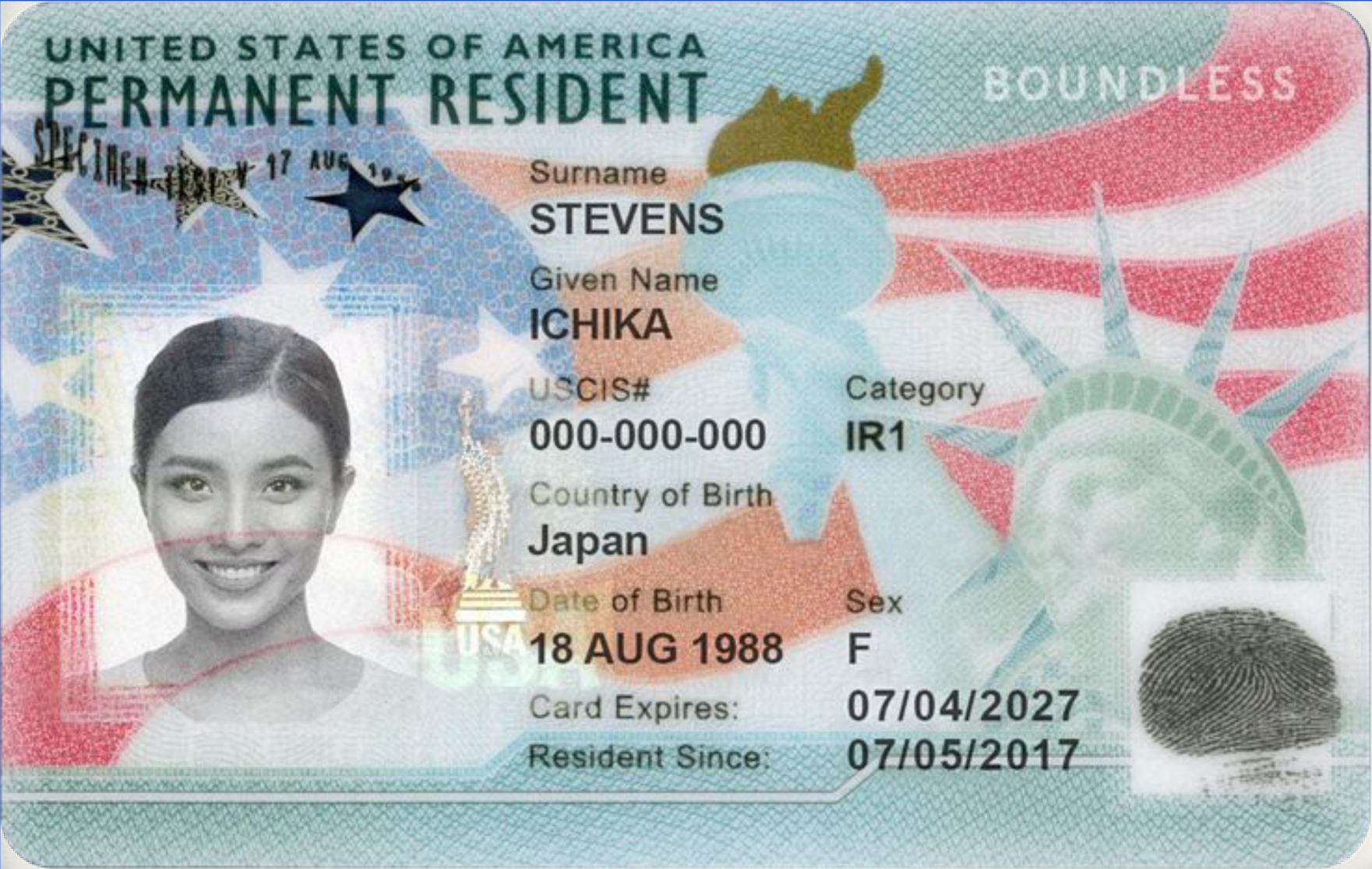}
  \caption{Image Case 4}
  \label{fig:understanding4} 
\end{figure}

\noindent 
\textcolor{blue}{CogVLM2:} \\
\{"Surname": "STEVENS", "Given Name": "ICHIKA", "USCIS \#": "000-000-000", "Category": "IR1", "Country of Birth": "Japan", "Date of Birth": "18 AUG 1988", "SEX": "F", "Card Expires": "07/04/2027", "Resident Since": "07/05/2017"\}
\end{promptbox}

\begin{promptbox}[Landmark Recognition]
\noindent
\textcolor{blue}{Prompt:} \\
Describe the landmark in the image.

\begin{figure}[H] %
  \centering
\includegraphics[width=\linewidth]{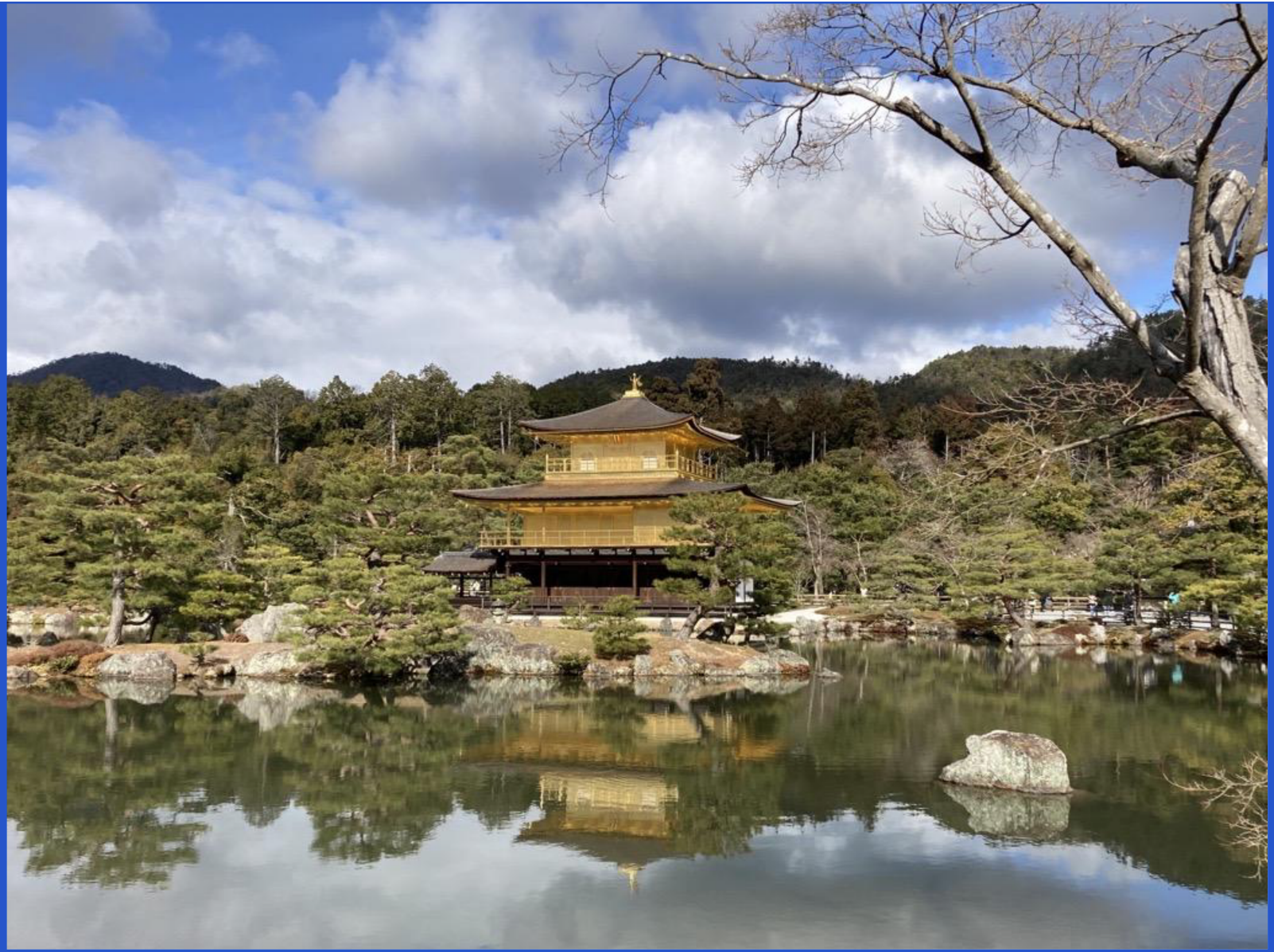}
  \caption{Image Case 5}
  \label{fig:understanding5} 
\end{figure}

\noindent 
\textcolor{blue}{CogVLM2:} \\
This image shows the Kinkaku-ji Temple, also known as the Golden Pavilion, located in Kyoto, Japan. The temple is a three-story Buddhist temple that is covered in gold leaf and sits on a pond, creating a stunning reflection on the water. The surrounding landscape is filled with trees and rocks, creating a peaceful and serene atmosphere. It is a popular tourist attraction and a designated National Special Historic Site.

\end{promptbox}

\subsection{Reasoning and Analysis}
\begin{promptbox}[Image Analysis]
\noindent
\textcolor{blue}{Prompt:} \\
\begin{CJK*}{UTF8}{gbsn}
请详细分析这张图片及其含义。
\end{CJK*}
\begin{figure}[H] %
  \centering
\includegraphics[width=0.8\linewidth]{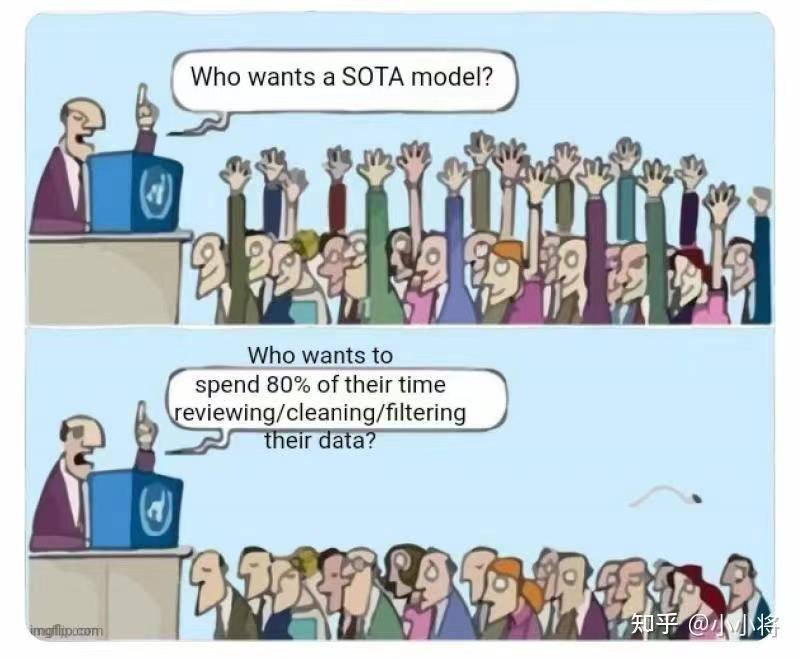}
  \caption{Image Case 6}
  \label{fig:reasonging1} 
\end{figure}

\noindent 
\textcolor{blue}{CogVLM2:} \\
\begin{CJK*}{UTF8}{gbsn}
这张图片展示了一个关于SOTA模型的投票场景。首先，有人提出“Who wants a SOTA model?”，然后大多数人都举手表示愿意。然而，接下来的问题是“Who wants to spend 80\% of their time reviewing/cleaning/filtering their data?”，只有少数人举手。图片的含义在于，虽然大多数人可能渴望使用最新的技术和模型，但他们可能不愿意投入大量的时间和精力来维护和准备数据，这可能是因为数据准备工作的复杂性和耗时性。

\end{CJK*}
\end{promptbox}

\begin{promptbox}[Image Analysis and World Knowledge]
\noindent
\textcolor{blue}{Prompt:} \\
\begin{CJK*}{UTF8}{gbsn}
哪个瓜最熟？给出理由

\end{CJK*}
\begin{figure}[H] %
  \centering
\includegraphics[width=0.6\linewidth]{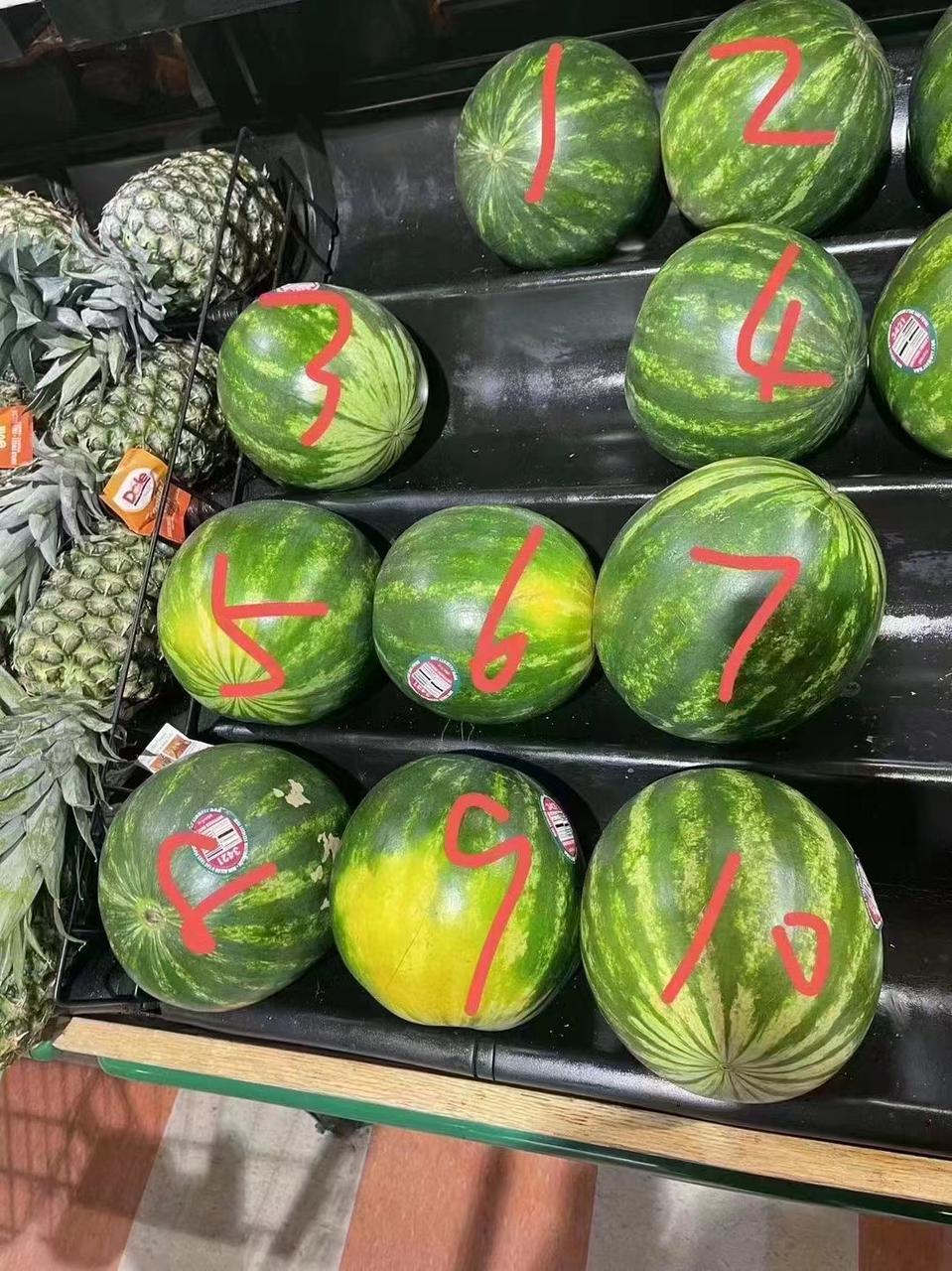}
  \caption{Image Case 7}
  \label{fig:reasonging2} 
\end{figure}

\noindent 
\textcolor{blue}{CogVLM2:} \\
\begin{CJK*}{UTF8}{gbsn}
根据图片，标号为9的瓜看起来最熟。它的底部有黄色，表皮上有明显的黑色斑点，这通常是成熟的迹象。

\end{CJK*}
\end{promptbox}

\begin{promptbox}[Image Analysis and World Knowledge]
\noindent
\textcolor{blue}{Prompt:} \\
Compare the average kinetic energies of the particles in each sample. The diagrams below show two pure samples of gas in identical closed, rigid containers. Each colored ball represents one gas particle. Both samples have the same number of particles. Which sample has the higher temperature? Sample A or sample B?

\begin{figure}[H] %
  \centering
\includegraphics[width=0.8\linewidth]{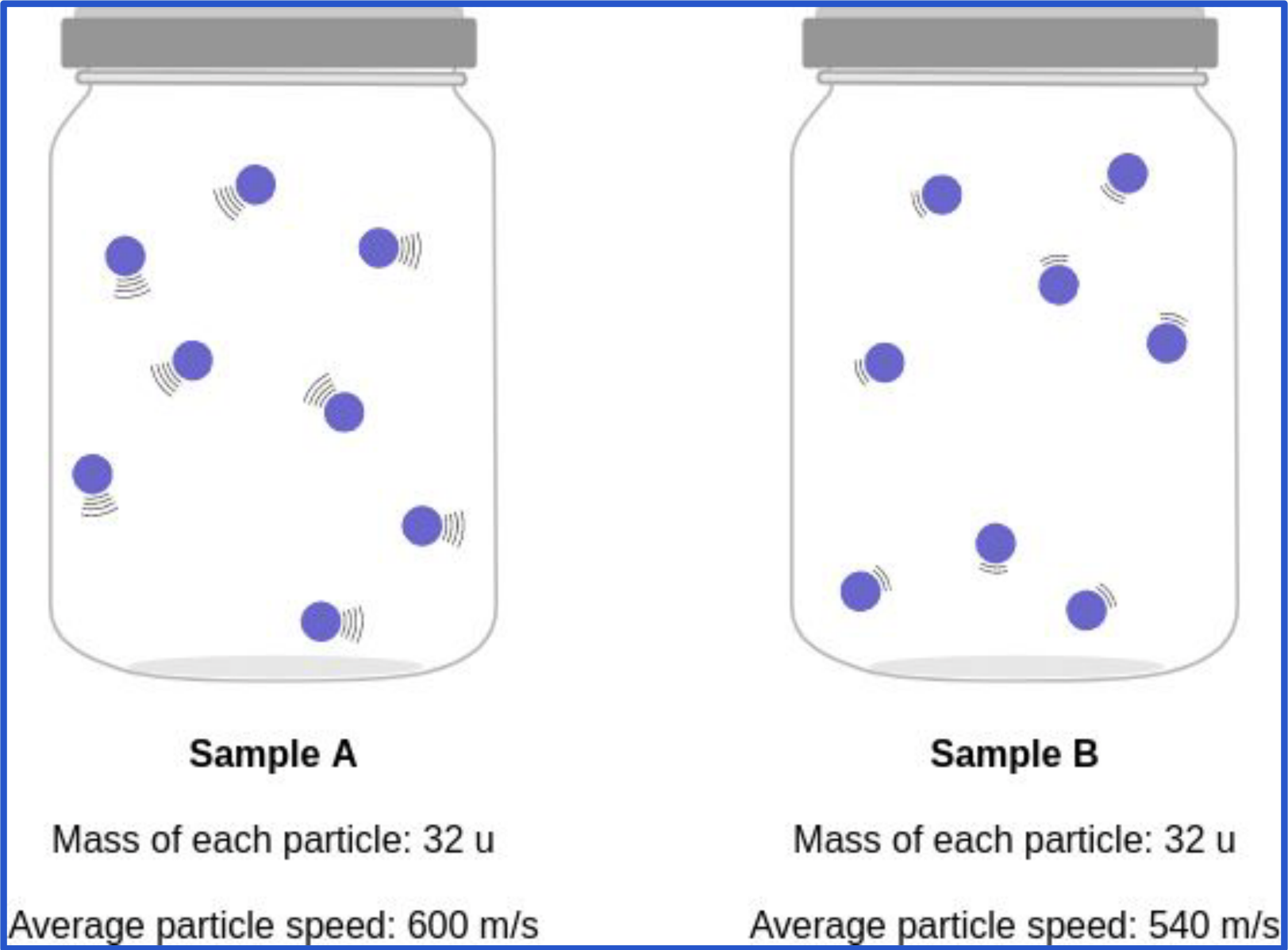}
  \caption{Image Case 8}
  \label{fig:reasonging3} 
\end{figure}

\noindent 
\textcolor{blue}{CogVLM2:} \\
To determine the temperature of a gas sample, we can use the formula for kinetic energy (KE), which is given by  $KE = 0.5 \times m \times v^2$, where m is the mass of the particle and v is its speed. Since both samples have particles with the same mass, the temperature will be determined by the average speed of the particles. Sample A has an average particle speed of 600 m/s, while Sample B has an average particle speed of 540 m/s. Therefore, Sample A has a higher average particle speed and, consequently, a higher average kinetic energy and temperature.

\end{promptbox}

\begin{promptbox}[Figure Analysis and Table generation]
\noindent
\textcolor{blue}{Prompt:} \\
The data point that stands out is the United States’ landfilled plastic waste, which is 73\%. This is significantly higher than the other countries and regions shown. This implies that the United States is not doing as much to recycle or incinerate its plastic waste as other countries and regions.

\begin{figure}[H] %
  \centering
\includegraphics[width=\linewidth]{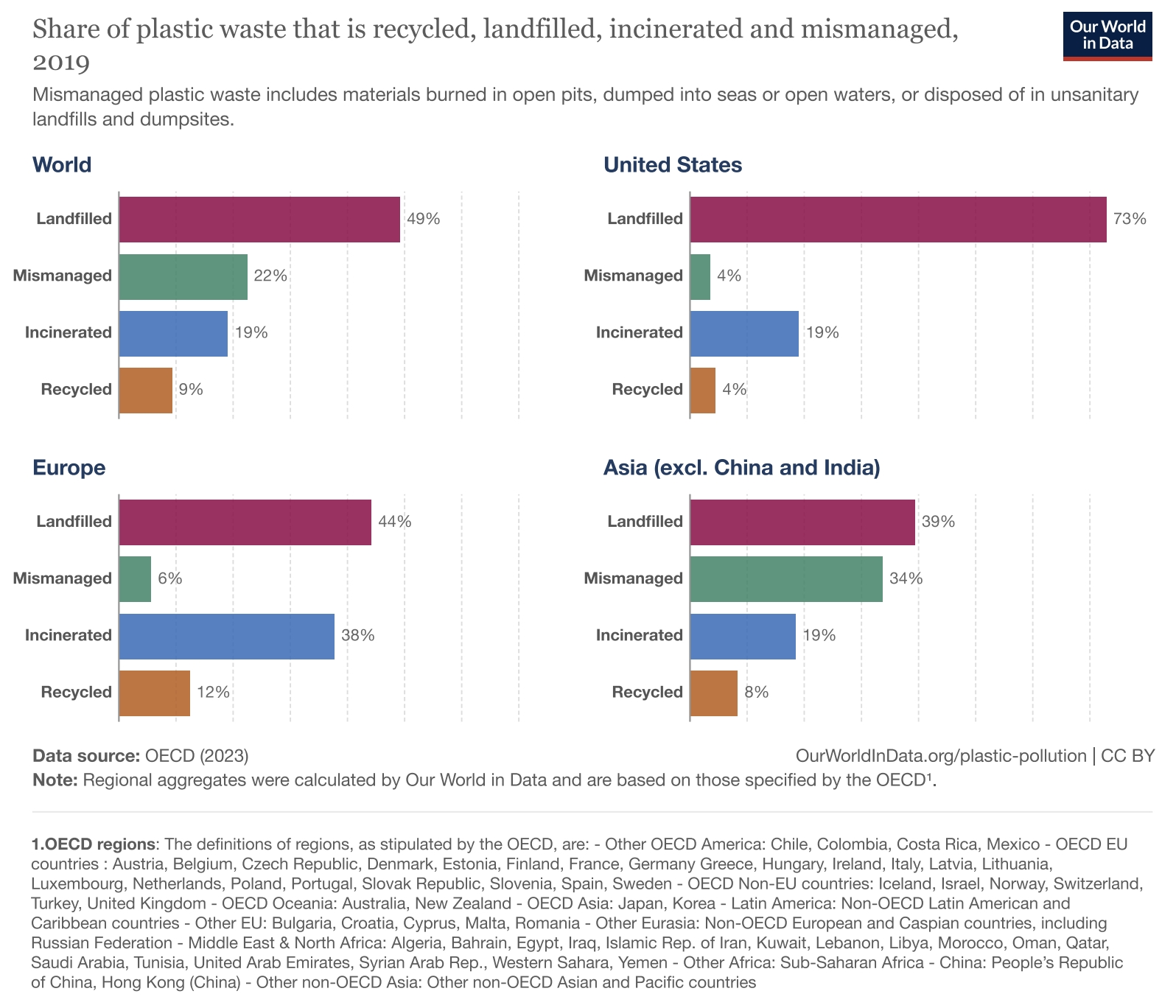}
  \caption{Image Case 9}
  \label{fig:reasonging4} 
\end{figure}

\noindent 
\textcolor{blue}{CogVLM2:} \\
One notable data point is the United States' high percentage of landfilled plastic waste at 73\%. This stands out as it is significantly higher than other regions, implying that the US has a substantial portion of its plastic waste that is not recycled, incinerated, or managed in other ways. Here's a markdown table for all the data shown: \\
\noindent 
| Region                    | Landfilled | Mismanaged | Incinerated | Recycled |\\
|---------------------------|------------|------------|------------|------------|\\
| World                     | 49\%        | 22\%        | 19\%         | 9\%       |\\
| United States             | 73\%        | 4\%         | 19\%         | 4\%       |\\
| Europe                    | 44\%        | 6\%         | 38\%         | 12\%      |\\
| Asia (excl. China and India) | 39\%      | 34\%        | 19\%         | 8\%       |

\end{promptbox}

\begin{promptbox}[Meme Understanding]
\noindent
\textcolor{blue}{Prompt:} \\
Can you explain this meme?

\begin{figure}[H] %
  \centering
\includegraphics[width=0.6\linewidth]{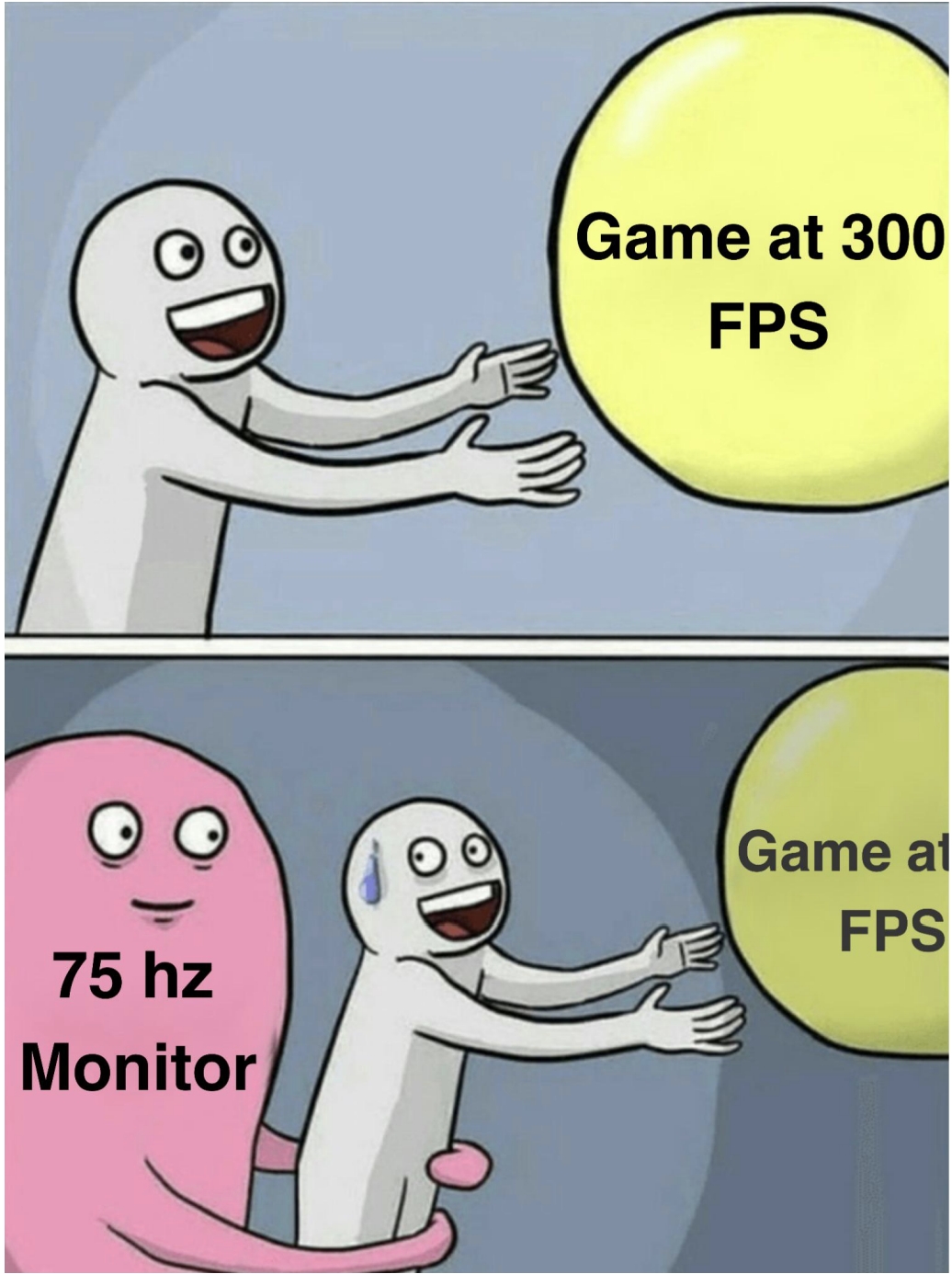}
  \caption{Image Case 10}
  \label{fig:reasonging5} 
\end{figure}

\noindent 
\textcolor{blue}{CogVLM2:} \\
Certainly! This meme is a humorous representation of the discrepancy between the ideal gaming experience and the limitations of one's hardware. The top panel shows a character excitedly reaching out to a balloon labeled 'Game at 300 FPS', symbolizing the desire to play a high frame rate game. The balloon is just out of reach, indicating that the character's current monitor, labeled '75 Hz Monitor', can't support such a high frame rate. The bottom panel shows the same character, now with sweat on their face, trying to reach the balloon with the help of another character labeled '75 Hz Monitor'. This character is depicted as being larger and more imposing, suggesting that even with their assistance, the challenge remains too great. The meme humorously highlights the frustration gamers feel when they want to enjoy high frame rate games but are limited by their monitor's refresh rate.

\end{promptbox}

\subsection{Video Understanding}

\begin{promptbox}[Video Description]
\noindent
\textcolor{blue}{Prompt:} \\
Describe this video in detail.
\begin{figure}[H] %
  \centering
\includegraphics[width=0.99\linewidth]{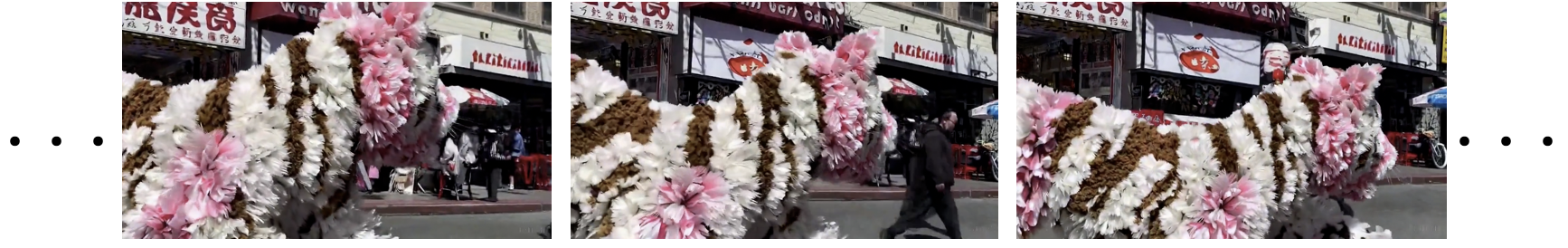}
  \caption{Video Case 1}
  \label{fig:video case 1} 
\end{figure}

\noindent 
\textcolor{blue}{CogVLM2-Video:} \\
The video features a large, colorful tiger sculpture made of various materials such as feathers and fabric. The sculpture is seen in different poses on an urban street lined with buildings featuring Asian signage.Throughout the video, people are present, some walking by and other standing near the sculptures or shops. At one point, there's a shower of pink petals that adds to the festive atmosphere. The scene captures a blend of traditional artistry and modern city life, creating a dynamic visual experience.

\end{promptbox}

\begin{promptbox}[Video Temporal Grounding]
\noindent
\textcolor{blue}{Prompt:} \\
What is the highlight of this video? At what second does it happen?
\begin{figure}[H] %
  \centering
\includegraphics[width=0.99\linewidth]{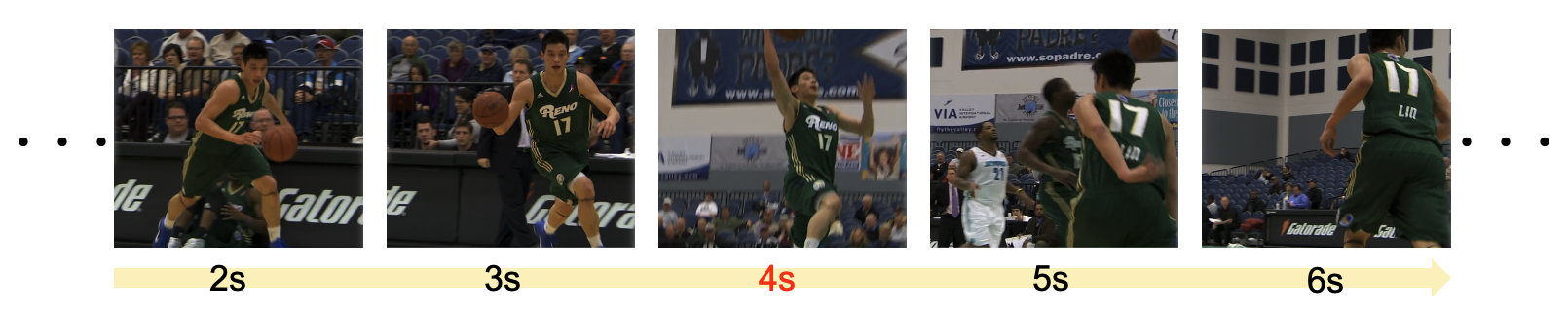}
  \caption{Video Case 2}
  \label{fig:video case 2} 
\end{figure}

\noindent 
\textcolor{blue}{CogVLM2-Video:} \\
The highlight of the video is when the player in the green jersey makes a basket. It happens at the 4th second.

\end{promptbox}

%% file: appendix/mvbench.tex
\section{MVBench Results}
\label{ap:mvbench}

We show the detailed results on MVBench as below:

\input{tables/mvbench}

%% file: tables/mvbench.tex
\begin{table}[htbp]
\tiny
\caption{\small Video understanding performance comparison on MVBench~\cite{li2023mvbench}. Best results are \textbf{bolded}.}
\label{mvbench}
\renewcommand{\arraystretch}{1.5}
\centering
\resizebox{\textwidth}{!}{%
\begin{tabular}{lccccccccccccccccccccc}
\toprule[1.25pt]
Model & AVG           & AA            & AC            & AL            & AP            & AS            & CO            & CI            & EN            & ER            & FA            & FP            & MA            & MC            & MD            & OE            & OI            & OS            & ST            & SC            & UA            \\\hline
IG-VLM GPT4V~\cite{kim2024imagegridworthvideo}              & 43.7          & 72.0          & 39.0          & 40.5          & 63.5          & 55.5          & 52.0          & 11.0          & 31.0          & 59.0          & 46.5          & 47.5          & 22.5          & 12.0          & 12.0          & 18.5          & 59.0          & 29.5          & 83.5          & 45.0          & 73.5          \\
ST-LLM~\cite{liu2023one}                    & 54.9          & 84.0          & 36.5          & 31.0          & 53.5          & 66.0          & 46.5          & 58.5          & 34.5          & 41.5          & 44.0          & 44.5          & 78.5          & 56.5          & 42.5          & 80.5          & 73.5          & 38.5          & 86.5          & 43.0          & 58.5          \\
ShareGPT4Video~\cite{chen2024sharegpt4video}            & 51.2          & 79.5          & 35.5          & 41.5          & 39.5          & 49.5          & 46.5          & 51.5          & 28.5          & 39.0          & 40.0          & 25.5          & 75.0          & 62.5          & 50.5          & 82.5          & 54.5          & 32.5          & 84.5          & 51.0          & 54.5          \\
VideoGPT+~\cite{maaz2024videogptintegratingimagevideo}                 & 58.7          & 83.0          & 39.5          & 34.0          & 60.0          & 69.0          & 50.0          & 60.0          & 29.5          & 44.0          & 48.5          & 53.0          & 90.5          & 71.0          & 44.0          & 85.5          & 75.5          & 36.0          & 89.5          & 45.0          & 66.5          \\
VideoChat2\_HD\_mistral~\cite{li2023mvbench}   & 62.3 & 79.5          & 60.0 & \textbf{87.5} & 50.0          & 68.5          & \textbf{93.5} & 71.5          & 36.5          & 45.0          & 49.5          & \textbf{87.0} & 40.0          & 76.0 & \textbf{92.0} & 53.0          & 62.0          & \textbf{45.5} & 36.0          & 44.0          & 69.5          \\
PLLaVA-34B~\cite{xu2024pllava}  & 58.1          & 82.0          & 40.5          & 49.5          & 53.0          & 67.5          & 66.5          & 59.0          & \textbf{39.5} & \textbf{63.5} & 47.0          & 50.0          & 70.0          & 43.0          & 37.5          & 68.5          & 67.5          & 36.5          & 91.0          & 51.5 & \textbf{79.0} \\
GPT-4o-2024-05-13~\cite{gpt4o} & 47.8 & 71.5          & 35.0          & 38.5          & 42.5          & 51.5          & 46.0          & 36.5          & 32.0          & 60.5 & 47.0 & 51.0          & 42.0          & 30.5          & 17.0          & 54.5          & 60.0          & 37.0          & 85.0          & 47.5          & 71.0  \\
Gemini 1.5 Pro~\cite{reid2024gemini}  & 52.6 & 55.0          & 32.0          & 50.0          & 56.0          & 66.0          & 51.0          & 46.5          & 29.0          & 62.0 & 41.5 & 59.0          & 56.5          & 41.5          & 34.5          & 57.0          & 73.0          & 41.0          & 84.0          & 44.0    &74.0  \\\hline
CogVLM2-Video (ours)             & 62.3 & 85.5 & 41.5          & 31.5          & 65.5 & \textbf{79.5} & 58.5          & 77.0 & 28.5          & 42.5          & \textbf{54.0} & 57.0          & 91.5 & 73.0          & 48.0          & 91.0 & 78.0 & 36.0          & \textbf{91.5} & 47.0          & 68.5 \\  
GLM-4V-Plus (ours)             & \textbf{71.2} & \textbf{94.0} & \textbf{71.0}          & 70.0          & \textbf{66.0} & \textbf{79.5} & 76.5          & \textbf{82.0} & 36.0          & 54.0          & 48.5 & 53.5          & \textbf{97.5} & \textbf{86.5}          & 55.5          & \textbf{95.5} & \textbf{80.0} & 37.5     & 91.0 & \textbf{77.5}          & 72.0 \\   
\bottomrule[1.25pt]
    \end{tabular}}
\end{table}

%% file: 0_main.bbl
\begin{thebibliography}{10}

\bibitem{TallyQA}
M.~Acharya, K.~Kafle, and C.~Kanan.
\newblock Tallyqa: Answering complex counting questions.
\newblock In {\em Proc. of Association for the Advancement of Artificial Intelligence}, 2019.

\bibitem{GPT4}
J.~Achiam, S.~Adler, S.~Agarwal, L.~Ahmad, I.~Akkaya, F.~L. Aleman, D.~Almeida, J.~Altenschmidt, S.~Altman, S.~Anadkat, et~al.
\newblock Gpt-4 technical report.
\newblock {\em arXiv preprint arXiv:2303.08774}, 2023.

\bibitem{alexey2020image}
D.~Alexey.
\newblock An image is worth 16x16 words: Transformers for image recognition at scale.
\newblock {\em arXiv preprint arXiv: 2010.11929}, 2020.

\bibitem{vqav2}
S.~Antol, A.~Agrawal, J.~Lu, M.~Mitchell, D.~Batra, C.~L. Zitnick, and D.~Parikh.
\newblock Vqa: Visual question answering.
\newblock In {\em Proc. of International Conference on Computer Vision}, pages 2425--2433, 2015.

\bibitem{bai2023qwen}
J.~Bai, S.~Bai, S.~Yang, S.~Wang, S.~Tan, P.~Wang, J.~Lin, C.~Zhou, and J.~Zhou.
\newblock Qwen-vl: A frontier large vision-language model with versatile abilities.
\newblock {\em arXiv preprint arXiv:2308.12966}, 2023.

\bibitem{Qwen-VL}
J.~Bai, S.~Bai, S.~Yang, S.~Wang, S.~Tan, P.~Wang, J.~Lin, C.~Zhou, and J.~Zhou.
\newblock Qwen-vl: A versatile vision-language model for understanding, localization, text reading, and beyond.
\newblock {\em arXiv preprint arXiv:2308.12966}, 2023.

\bibitem{STVQA}
A.~F. Biten, R.~Tito, A.~Mafla, L.~Gomez, M.~Rusiñol, C.~Jawahar, E.~Valveny, and D.~Karatzas.
\newblock Scene text visual question answering.
\newblock In {\em Proc. of International Conference on Computer Vision}, pages 4290--4300, 2019.

\bibitem{blecher2023nougat}
L.~Blecher, G.~Cucurull, T.~Scialom, and R.~Stojnic.
\newblock Nougat: Neural optical understanding for academic documents.
\newblock {\em arXiv preprint arXiv:2308.13418}, 2023.

\bibitem{kakaobrain2022coyo-700m}
M.~Byeon, B.~Park, H.~Kim, S.~Lee, W.~Baek, and S.~Kim.
\newblock Coyo-700m: Image-text pair dataset.
\newblock \url{https://github.com/kakaobrain/coyo-dataset}, 2022.

\bibitem{GeoQA}
J.~Chen, J.~Tang, J.~Qin, X.~Liang, L.~Liu, E.~P. Xing, and L.~Lin.
\newblock Geoqa: A geometric question answering benchmark towards multimodal numerical reasoning, 2022.

\bibitem{chen2024we}
L.~Chen, J.~Li, X.~Dong, P.~Zhang, Y.~Zang, Z.~Chen, H.~Duan, J.~Wang, Y.~Qiao, D.~Lin, et~al.
\newblock Are we on the right way for evaluating large vision-language models?
\newblock {\em arXiv preprint arXiv:2403.20330}, 2024.

\bibitem{chen2024sharegpt4video}
L.~Chen, X.~Wei, J.~Li, X.~Dong, P.~Zhang, Y.~Zang, Z.~Chen, H.~Duan, B.~Lin, Z.~Tang, L.~Yuan, Y.~Qiao, D.~Lin, F.~Zhao, and J.~Wang.
\newblock Sharegpt4video: Improving video understanding and generation with better captions.
\newblock {\em arXiv preprint arXiv:2406.04325}, 2024.

\bibitem{chen2022pali}
X.~Chen, X.~Wang, S.~Changpinyo, A.~Piergiovanni, P.~Padlewski, D.~Salz, S.~Goodman, A.~Grycner, B.~Mustafa, L.~Beyer, et~al.
\newblock Pali: A jointly-scaled multilingual language-image model.
\newblock {\em arXiv preprint arXiv:2209.06794}, 2022.

\bibitem{chen2023internvl}
Z.~Chen, J.~Wu, W.~Wang, W.~Su, G.~Chen, S.~Xing, M.~Zhong, Q.~Zhang, X.~Zhu, L.~Lu, B.~Li, P.~Luo, T.~Lu, Y.~Qiao, and J.~Dai.
\newblock Internvl: Scaling up vision foundation models and aligning for generic visual-linguistic tasks.
\newblock {\em arXiv preprint arXiv:2312.14238}, 2023.

\bibitem{gadre2024datacomp}
S.~Y. Gadre, G.~Ilharco, A.~Fang, J.~Hayase, G.~Smyrnis, T.~Nguyen, R.~Marten, M.~Wortsman, D.~Ghosh, J.~Zhang, et~al.
\newblock Datacomp: In search of the next generation of multimodal datasets.
\newblock {\em Advances in Neural Information Processing Systems}, 36, 2024.

\bibitem{Geo170K}
J.~Gao, R.~Pi, J.~Zhang, J.~Ye, W.~Zhong, Y.~Wang, L.~Hong, J.~Han, H.~Xu, Z.~Li, et~al.
\newblock G-llava: Solving geometric problem with multi-modal large language model.
\newblock {\em arXiv preprint arXiv:2312.11370}, 2023.

\bibitem{glm2024chatglm}
T.~GLM, A.~Zeng, B.~Xu, B.~Wang, C.~Zhang, D.~Yin, D.~Rojas, G.~Feng, H.~Zhao, H.~Lai, et~al.
\newblock Chatglm: A family of large language models from glm-130b to glm-4 all tools.
\newblock {\em arXiv preprint arXiv:2406.12793}, 2024.

\bibitem{sth}
R.~Goyal, S.~E. Kahou, V.~Michalski, J.~Materzynska, S.~Westphal, H.~Kim, V.~Haenel, I.~Fr{\"u}nd, P.~Yianilos, M.~Mueller-Freitag, F.~Hoppe, C.~Thurau, I.~Bax, and R.~Memisevic.
\newblock The “something something” video database for learning and evaluating visual common sense.
\newblock In {\em Proc. of International Conference on Computer Vision}, 2017.

\bibitem{ego4d}
K.~Grauman, A.~Westbury, E.~Byrne, Z.~Chavis, A.~Furnari, R.~Girdhar, J.~Hamburger, H.~Jiang, M.~Liu, X.~Liu, M.~Martin, T.~Nagarajan, I.~Radosavovic, S.~K. Ramakrishnan, F.~Ryan, J.~Sharma, M.~Wray, M.~Xu, E.~Z. Xu, C.~Zhao, S.~Bansal, D.~Batra, V.~Cartillier, S.~Crane, T.~Do, M.~Doulaty, A.~Erapalli, C.~Feichtenhofer, A.~Fragomeni, Q.~Fu, C.~Fuegen, A.~Gebreselasie, C.~Gonz{\'a}lez, J.~M. Hillis, X.~Huang, Y.~Huang, W.~Jia, W.~Khoo, J.~Kol{\'a}r, S.~Kottur, A.~Kumar, F.~Landini, C.~Li, Y.~Li, Z.~Li, K.~Mangalam, R.~Modhugu, J.~Munro, T.~Murrell, T.~Nishiyasu, W.~Price, P.~R. Puentes, M.~Ramazanova, L.~Sari, K.~K. Somasundaram, A.~Southerland, Y.~Sugano, R.~Tao, M.~Vo, Y.~Wang, X.~Wu, T.~Yagi, Y.~Zhu, P.~Arbel{\'a}ez, D.~J. Crandall, D.~Damen, G.~M. Farinella, B.~Ghanem, V.~K. Ithapu, C.~V. Jawahar, H.~Joo, K.~Kitani, H.~Li, R.~A. Newcombe, A.~Oliva, H.~S. Park, J.~M. Rehg, Y.~Sato, J.~Shi, M.~Z. Shou, A.~Torralba, L.~Torresani, M.~Yan, and J.~Malik.
\newblock Ego4d: Around the world in 3,000 hours of egocentric video.
\newblock In {\em Proc. of Computer Vision and Pattern Recognition}, 2022.

\bibitem{hong2024cogagent}
W.~Hong, W.~Wang, Q.~Lv, J.~Xu, W.~Yu, J.~Ji, Y.~Wang, Z.~Wang, Y.~Dong, M.~Ding, et~al.
\newblock Cogagent: A visual language model for gui agents.
\newblock In {\em Proceedings of the IEEE/CVF Conference on Computer Vision and Pattern Recognition}, pages 14281--14290, 2024.

\bibitem{tgif_qa}
Y.~Jang, Y.~Song, Y.~Yu, Y.~Kim, and G.~Kim.
\newblock Tgif-qa: Toward spatio-temporal reasoning in visual question answering.
\newblock In {\em Proc. of Computer Vision and Pattern Recognition}, 2017.

\bibitem{DVQA}
K.~Kafle, S.~Cohen, B.~Price, and C.~Kanan.
\newblock Dvqa: Understanding data visualizations via question answering.
\newblock In {\em Proc. of Computer Vision and Pattern Recognition}, 2018.

\bibitem{TDIUC}
K.~Kafle and C.~Kanan.
\newblock An analysis of visual question answering algorithms.
\newblock In {\em Proceedings of the IEEE international conference on computer vision}, pages 1965--1973, 2017.

\bibitem{FigureQA}
S.~E. Kahou, V.~Michalski, A.~Atkinson, A.~Kadar, A.~Trischler, and Y.~Bengio.
\newblock Figureqa: An annotated figure dataset for visual reasoning, 2018.

\bibitem{Geomverse}
M.~Kazemi, H.~Alvari, A.~Anand, J.~Wu, X.~Chen, and R.~Soricut.
\newblock Geomverse: A systematic evaluation of large models for geometric reasoning.
\newblock {\em arXiv preprint arXiv:2312.12241}, 2023.

\bibitem{Ai2D}
A.~Kembhavi, M.~Salvato, E.~Kolve, M.~Seo, H.~Hajishirzi, and A.~Farhadi.
\newblock A diagram is worth a dozen images, 2016.

\bibitem{kembhavi2016diagram}
A.~Kembhavi, M.~Salvato, E.~Kolve, M.~Seo, H.~Hajishirzi, and A.~Farhadi.
\newblock A diagram is worth a dozen images.
\newblock In {\em Computer Vision--ECCV 2016: 14th European Conference, Amsterdam, The Netherlands, October 11--14, 2016, Proceedings, Part IV 14}, pages 235--251. Springer, 2016.

\bibitem{TQA}
A.~Kembhavi, M.~Seo, D.~Schwenk, J.~Choi, A.~Farhadi, and H.~Hajishirzi.
\newblock Are you smarter than a sixth grader? textbook question answering for multimodal machine comprehension.
\newblock In {\em Proc. of Computer Vision and Pattern Recognition}, pages 5376--5384, 2017.

\bibitem{kim2024imagegridworthvideo}
W.~Kim, C.~Choi, W.~Lee, and W.~Rhee.
\newblock An image grid can be worth a video: Zero-shot video question answering using a vlm, 2024.

\bibitem{VGQA}
R.~Krishna, Y.~Zhu, O.~Groth, J.~Johnson, K.~Hata, J.~Kravitz, S.~Chen, Y.~Kalantidis, L.-J. Li, D.~A. Shamma, et~al.
\newblock Visual genome: Connecting language and vision using crowdsourced dense image annotations.
\newblock {\em International journal of computer vision}, 123:32--73, 2017.

\bibitem{VQARAD}
J.~J. Lau, S.~Gayen, A.~Ben~Abacha, and D.~Demner-Fushman.
\newblock A dataset of clinically generated visual questions and answers about radiology images.
\newblock {\em Scientific data}, 5(1):1--10, 2018.

\bibitem{li2022pp}
C.~Li, W.~Liu, R.~Guo, X.~Yin, K.~Jiang, Y.~Du, Y.~Du, L.~Zhu, B.~Lai, X.~Hu, et~al.
\newblock Pp-ocrv3: More attempts for the improvement of ultra lightweight ocr system.
\newblock {\em arXiv preprint arXiv:2206.03001}, 2022.

\bibitem{li2023blip}
J.~Li, D.~Li, S.~Savarese, and S.~Hoi.
\newblock Blip-2: Bootstrapping language-image pre-training with frozen image encoders and large language models.
\newblock In {\em International conference on machine learning}, pages 19730--19742. PMLR, 2023.

\bibitem{videochat}
K.~Li, Y.~He, Y.~Wang, Y.~Li, W.~Wang, P.~Luo, Y.~Wang, L.~Wang, and Y.~Qiao.
\newblock Videochat: Chat-centric video understanding.
\newblock {\em ArXiv}, abs/2305.06355, 2023.

\bibitem{li2023mvbench}
K.~Li, Y.~Wang, Y.~He, Y.~Li, Y.~Wang, Y.~Liu, Z.~Wang, J.~Xu, G.~Chen, P.~Luo, L.~Wang, and Y.~Qiao.
\newblock {MVBench}: {A} comprehensive multi-modal video understanding benchmark, 2023.

\bibitem{uniformerv2}
K.~Li, Y.~Wang, Y.~He, Y.~Li, Y.~Wang, L.~Wang, and Y.~Qiao.
\newblock Uniformerv2: Spatiotemporal learning by arming image vits with video uniformer.
\newblock {\em ArXiv}, abs/2211.09552, 2022.

\bibitem{ArxivQA}
L.~Li, Y.~Wang, R.~Xu, P.~Wang, X.~Feng, L.~Kong, and Q.~Liu.
\newblock Multimodal arxiv: A dataset for improving scientific comprehension of large vision-language models, 2024.

\bibitem{li2024minigemini}
Y.~Li, Y.~Zhang, C.~Wang, Z.~Zhong, Y.~Chen, R.~Chu, S.~Liu, and J.~Jia.
\newblock Mini-gemini: Mining the potential of multi-modality vision language models.
\newblock {\em arXiv preprint arXiv:2403.18814}, 2024.

\bibitem{lin2023univtg}
K.~Q. Lin, P.~Zhang, J.~Chen, S.~Pramanick, D.~Gao, A.~J. Wang, R.~Yan, and M.~Z. Shou.
\newblock Univtg: Towards unified video-language temporal grounding.
\newblock In {\em Proc. of International Conference on Computer Vision}, 2023.

\bibitem{VSR}
F.~Liu, G.~Emerson, and N.~Collier.
\newblock Visual spatial reasoning.
\newblock {\em Proc. of IEEE International Conference on Automatic Face and Gesture Recognition}, 11, 2023.

\bibitem{liu2024improvedbaselinesvisualinstruction}
H.~Liu, C.~Li, Y.~Li, and Y.~J. Lee.
\newblock Improved baselines with visual instruction tuning, 2024.

\bibitem{liu2024llavanext}
H.~Liu, C.~Li, Y.~Li, B.~Li, Y.~Zhang, S.~Shen, and Y.~J. Lee.
\newblock Llava-next: Improved reasoning, ocr, and world knowledge, January 2024.

\bibitem{liu2024visual}
H.~Liu, C.~Li, Q.~Wu, and Y.~J. Lee.
\newblock Visual instruction tuning.
\newblock {\em Proc. of Neural Information Processing Systems}, 36, 2024.

\bibitem{liu2023one}
R.~Liu, C.~Li, H.~Tang, Y.~Ge, Y.~Shan, and G.~Li.
\newblock St-llm: Large language models are effective temporal learners.
\newblock {\em https://arxiv.org/abs/2404.00308}, 2023.

\bibitem{MMBench}
Y.~Liu, H.~Duan, Y.~Zhang, B.~Li, S.~Zhang, W.~Zhao, Y.~Yuan, J.~Wang, C.~He, Z.~Liu, K.~Chen, and D.~Lin.
\newblock Mmbench: Is your multi-modal model an all-around player?
\newblock {\em arXiv:2307.06281}, 2023.

\bibitem{liu2024hiddenmysteryocrlarge}
Y.~Liu, Z.~Li, B.~Yang, C.~Li, X.~Yin, C.~lin Liu, L.~Jin, and X.~Bai.
\newblock On the hidden mystery of ocr in large multimodal models, 2024.

\bibitem{InterGPS}
P.~Lu, R.~Gong, S.~Jiang, L.~Qiu, S.~Huang, X.~Liang, and S.-C. Zhu.
\newblock Inter-gps: Interpretable geometry problem solving with formal language and symbolic reasoning.
\newblock In {\em The Joint Conference of the 59th Annual Meeting of the Association for Computational Linguistics and the 11th International Joint Conference on Natural Language Processing (ACL-IJCNLP 2021)}, 2021.

\bibitem{scienceqa}
P.~Lu, S.~Mishra, T.~Xia, L.~Qiu, K.-W. Chang, S.-C. Zhu, O.~Tafjord, P.~Clark, and A.~Kalyan.
\newblock Learn to explain: Multimodal reasoning via thought chains for science question answering.
\newblock {\em Proc. of Neural Information Processing Systems}, 35:2507--2521, 2022.

\bibitem{TabMWP}
P.~Lu, L.~Qiu, K.-W. Chang, Y.~N. Wu, S.-C. Zhu, T.~Rajpurohit, P.~Clark, and A.~Kalyan.
\newblock Dynamic prompt learning via policy gradient for semi-structured mathematical reasoning.
\newblock In {\em Proc. of International Conference on Learning Representations}, 2023.

\bibitem{IconQA}
P.~Lu, L.~Qiu, J.~Chen, T.~Xia, Y.~Zhao, W.~Zhang, Z.~Yu, X.~Liang, and S.-C. Zhu.
\newblock Iconqa: A new benchmark for abstract diagram understanding and visual language reasoning.
\newblock In {\em The 35th Conference on Neural Information Processing Systems (NeurIPS) Track on Datasets and Benchmarks}, 2021.

\bibitem{luo2023towards}
D.~Luo, J.~Huang, S.~Gong, H.~Jin, and Y.~Liu.
\newblock Towards generalisable video moment retrieval: Visual-dynamic injection to image-text pre-training.
\newblock In {\em Proc. of Computer Vision and Pattern Recognition}, 2023.

\bibitem{maaz2024videogptintegratingimagevideo}
M.~Maaz, H.~Rasheed, S.~Khan, and F.~Khan.
\newblock Videogpt+: Integrating image and video encoders for enhanced video understanding, 2024.

\bibitem{maaz2023videovhatgpt}
M.~Maaz, H.~Rasheed, S.~Khan, and F.~S. Khan.
\newblock Video-chatgpt: Towards detailed video understanding via large vision and language models.
\newblock In {\em Proc. of IEEE International Conference on Automatic Face and Gesture Recognition}, 2024.

\bibitem{Maaz2023VideoChatGPT}
M.~Maaz, H.~Rasheed, S.~Khan, and F.~S. Khan.
\newblock Video-chatgpt: Towards detailed video understanding via large vision and language models.
\newblock In {\em Proceedings of the 62nd Annual Meeting of the Association for Computational Linguistics (ACL 2024)}, 2024.

\bibitem{okvqa}
K.~Marino, M.~Rastegari, A.~Farhadi, and R.~Mottaghi.
\newblock Ok-vqa: A visual question answering benchmark requiring external knowledge.
\newblock In {\em Proc. of Computer Vision and Pattern Recognition}, pages 3195--3204, 2019.

\bibitem{IAM}
U.-V. Marti and H.~Bunke.
\newblock The iam-database: An english sentence database for offline handwriting recognition.
\newblock {\em International Journal on Document Analysis and Recognition}, 5:39--46, 11 2002.

\bibitem{ChartQA}
A.~Masry, D.~Long, J.~Q. Tan, S.~Joty, and E.~Hoque.
\newblock {C}hart{QA}: A benchmark for question answering about charts with visual and logical reasoning.
\newblock In {\em Proc. of IEEE International Conference on Automatic Face and Gesture Recognition}, pages 2263--2279, Dublin, Ireland, May 2022. Association for Computational Linguistics.

\bibitem{masry2022chartqa}
A.~Masry, D.~X. Long, J.~Q. Tan, S.~Joty, and E.~Hoque.
\newblock Chartqa: A benchmark for question answering about charts with visual and logical reasoning.
\newblock {\em arXiv preprint arXiv:2203.10244}, 2022.

\bibitem{masry2022chartqabenchmarkquestionanswering}
A.~Masry, D.~X. Long, J.~Q. Tan, S.~Joty, and E.~Hoque.
\newblock Chartqa: A benchmark for question answering about charts with visual and logical reasoning, 2022.

\bibitem{InfographicVQA}
M.~Mathew, V.~Bagal, R.~Tito, D.~Karatzas, E.~Valveny, and C.~V. Jawahar.
\newblock Infographicvqa.
\newblock In {\em Proc. of IEEE Winter Conference on Applications of Computer Vision}, pages 2582--2591, 2022.

\bibitem{mathew2021docvqa}
M.~Mathew, D.~Karatzas, and C.~Jawahar.
\newblock Docvqa: A dataset for vqa on document images.
\newblock In {\em Proceedings of the IEEE/CVF winter conference on applications of computer vision}, pages 2200--2209, 2021.

\bibitem{DocVQA}
M.~Mathew, D.~Karatzas, and C.~V. Jawahar.
\newblock Docvqa: A dataset for vqa on document images.
\newblock In {\em Proc. of IEEE Winter Conference on Applications of Computer Vision}, pages 2199--2208, 2021.

\bibitem{mathew2021docvqadatasetvqadocument}
M.~Mathew, D.~Karatzas, and C.~V. Jawahar.
\newblock Docvqa: A dataset for vqa on document images, 2021.

\bibitem{ocrvqa}
A.~Mishra, S.~Shekhar, A.~K. Singh, and A.~Chakraborty.
\newblock Ocr-vqa: Visual question answering by reading text in images.
\newblock In {\em 2019 international conference on document analysis and recognition (ICDAR)}, pages 947--952. IEEE, 2019.

\bibitem{gpt4o}
OpenAI.
\newblock Gpt-4o.
\newblock 2024.

\bibitem{reid2024gemini}
M.~Reid, N.~Savinov, D.~Teplyashin, D.~Lepikhin, T.~Lillicrap, J.-b. Alayrac, R.~Soricut, A.~Lazaridou, O.~Firat, J.~Schrittwieser, et~al.
\newblock Gemini 1.5: Unlocking multimodal understanding across millions of tokens of context.
\newblock {\em arXiv preprint arXiv:2403.05530}, 2024.

\bibitem{ren2023timechat}
S.~Ren, L.~Yao, S.~Li, X.~Sun, and L.~Hou.
\newblock Timechat: A time-sensitive multimodal large language model for long video understanding.
\newblock {\em arXiv preprint arXiv:2312.02051}, 2023.

\bibitem{schuhmann2022laion}
C.~Schuhmann, R.~Beaumont, R.~Vencu, C.~Gordon, R.~Wightman, M.~Cherti, T.~Coombes, A.~Katta, C.~Mullis, M.~Wortsman, et~al.
\newblock Laion-5b: An open large-scale dataset for training next generation image-text models.
\newblock {\em Advances in Neural Information Processing Systems}, 35:25278--25294, 2022.

\bibitem{schuhmann2021laion}
C.~Schuhmann, R.~Vencu, R.~Beaumont, R.~Kaczmarczyk, C.~Mullis, A.~Katta, T.~Coombes, J.~Jitsev, and A.~Komatsuzaki.
\newblock Laion-400m: Open dataset of clip-filtered 400 million image-text pairs.
\newblock {\em arXiv preprint arXiv:2111.02114}, 2021.

\bibitem{A-OKVQA}
D.~Schwenk, A.~Khandelwal, C.~Clark, K.~Marino, and R.~Mottaghi.
\newblock A-okvqa: A benchmark for visual question answering using world knowledge.
\newblock In {\em Proc. of European Conference on Computer Vision}, page 146–162, 2022.

\bibitem{shazeer2020glu}
N.~Shazeer.
\newblock Glu variants improve transformer, 2020.

\bibitem{singh2019towards}
A.~Singh, V.~Natarajan, M.~Shah, Y.~Jiang, X.~Chen, D.~Batra, D.~Parikh, and M.~Rohrbach.
\newblock Towards vqa models that can read.
\newblock In {\em Proceedings of the IEEE/CVF conference on computer vision and pattern recognition}, pages 8317--8326, 2019.

\bibitem{textvqa}
A.~Singh, V.~Natarajan, M.~Shah, Y.~Jiang, X.~Chen, D.~Batra, D.~Parikh, and M.~Rohrbach.
\newblock Towards vqa models that can read.
\newblock In {\em Proc. of Computer Vision and Pattern Recognition}, pages 8317--8326, 2019.

\bibitem{sun2023eva}
Q.~Sun, Y.~Fang, L.~Wu, X.~Wang, and Y.~Cao.
\newblock Eva-clip: Improved training techniques for clip at scale.
\newblock {\em arXiv preprint arXiv:2303.15389}, 2023.

\bibitem{VisText}
B.~J. Tang, A.~Boggust, and A.~Satyanarayan.
\newblock {VisText: A Benchmark for Semantically Rich Chart Captioning}.
\newblock In {\em Proc. of IEEE International Conference on Automatic Face and Gesture Recognition}, 2023.

\bibitem{tong2024cambrian1}
S.~Tong, E.~Brown, P.~Wu, S.~Woo, M.~Middepogu, S.~C. Akula, J.~Yang, S.~Yang, A.~Iyer, X.~Pan, A.~Wang, R.~Fergus, Y.~LeCun, and S.~Xie.
\newblock Cambrian-1: A fully open, vision-centric exploration of multimodal llms, 2024.

\bibitem{wang2024lvbench}
W.~Wang, Z.~He, W.~Hong, Y.~Cheng, X.~Zhang, J.~Qi, S.~Huang, B.~Xu, Y.~Dong, M.~Ding, et~al.
\newblock Lvbench: An extreme long video understanding benchmark.
\newblock {\em arXiv preprint arXiv:2406.08035}, 2024.

\bibitem{wang2023cogvlm}
W.~Wang, Q.~Lv, W.~Yu, W.~Hong, J.~Qi, Y.~Wang, J.~Ji, Z.~Yang, L.~Zhao, X.~Song, et~al.
\newblock Cogvlm: Visual expert for pretrained language models.
\newblock {\em arXiv preprint arXiv:2311.03079}, 2023.

\bibitem{wang2022negative}
Z.~Wang, L.~Wang, T.~Wu, T.~Li, and G.~Wu.
\newblock Negative sample matters: A renaissance of metric learning for temporal grounding.
\newblock In {\em Proc. of Association for the Advancement of Artificial Intelligence}, 2022.

\bibitem{nextqa}
J.~Xiao, X.~Shang, A.~Yao, and T.~seng Chua.
\newblock Next-qa: Next phase of question-answering to explaining temporal actions.
\newblock In {\em Proc. of Computer Vision and Pattern Recognition}, 2021.

\bibitem{xu2024pllava}
L.~Xu, Y.~Zhao, D.~Zhou, Z.~Lin, S.~K. Ng, and J.~Feng.
\newblock {PLLaVA}: Parameter-free {LLaVA} extension from images to videos for video dense captioning, 2024.

\bibitem{just_ask}
A.~Yang, A.~Miech, J.~Sivic, I.~Laptev, and C.~Schmid.
\newblock Just ask: Learning to answer questions from millions of narrated videos.
\newblock In {\em Proc. of International Conference on Computer Vision}, 2021.

\bibitem{yao2024minicpm}
Y.~Yao, T.~Yu, A.~Zhang, C.~Wang, J.~Cui, H.~Zhu, T.~Cai, H.~Li, W.~Zhao, Z.~He, et~al.
\newblock Minicpm-v: A gpt-4v level mllm on your phone.
\newblock {\em arXiv preprint arXiv:2408.01800}, 2024.

\bibitem{clevrer}
K.~Yi, C.~Gan, Y.~Li, P.~Kohli, J.~Wu, A.~Torralba, and J.~B. Tenenbaum.
\newblock Clevrer: Collision events for video representation and reasoning.
\newblock In {\em Proc. of International Conference on Learning Representations}, 2020.

\bibitem{yin2023survey}
S.~Yin, C.~Fu, S.~Zhao, K.~Li, X.~Sun, T.~Xu, and E.~Chen.
\newblock A survey on multimodal large language models.
\newblock {\em arXiv preprint arXiv:2306.13549}, 2023.

\bibitem{yu2023mm}
W.~Yu, Z.~Yang, L.~Li, J.~Wang, K.~Lin, Z.~Liu, X.~Wang, and L.~Wang.
\newblock Mm-vet: Evaluating large multimodal models for integrated capabilities.
\newblock {\em arXiv preprint arXiv:2308.02490}, 2023.

\bibitem{yu2024mm}
W.~Yu, Z.~Yang, L.~Li, J.~Wang, K.~Lin, Z.~Liu, X.~Wang, and L.~Wang.
\newblock Mm-vet: Evaluating large multimodal models for integrated capabilities.
\newblock In {\em Proc. of International Conference on Machine Learning}, 2024.

\bibitem{yue2023mmmu}
X.~Yue, Y.~Ni, K.~Zhang, T.~Zheng, R.~Liu, G.~Zhang, S.~Stevens, D.~Jiang, W.~Ren, Y.~Sun, C.~Wei, B.~Yu, R.~Yuan, R.~Sun, M.~Yin, B.~Zheng, Z.~Yang, Y.~Liu, W.~Huang, H.~Sun, Y.~Su, and W.~Chen.
\newblock Mmmu: A massive multi-discipline multimodal understanding and reasoning benchmark for expert agi.
\newblock In {\em Proc. of Computer Vision and Pattern Recognition}, 2024.

\bibitem{Raven}
C.~Zhang, F.~Gao, B.~Jia, Y.~Zhu, and S.-C. Zhu.
\newblock Raven: A dataset for relational and analogical visual reasoning.
\newblock In {\em Proceedings of the IEEE Conference on Computer Vision and Pattern Recognition (CVPR)}, 2019.

\bibitem{zhang2023video}
H.~Zhang, X.~Li, and L.~Bing.
\newblock Video-llama: An instruction-tuned audio-visual language model for video understanding.
\newblock {\em arXiv preprint arXiv:2306.02858}, 2023.

\bibitem{zhang2022glipv2}
H.~Zhang, P.~Zhang, X.~Hu, Y.-C. Chen, L.~Li, X.~Dai, L.~Wang, L.~Yuan, J.-N. Hwang, and J.~Gao.
\newblock Glipv2: Unifying localization and vision-language understanding.
\newblock {\em Proc. of Neural Information Processing Systems}, 35:36067--36080, 2022.

\bibitem{zhang2024vcr}
T.~Zhang, S.~Wang, L.~Li, G.~Zhang, P.~Taslakian, S.~Rajeswar, J.~Fu, B.~Liu, and Y.~Bengio.
\newblock Vcr: Visual caption restoration.
\newblock {\em arXiv preprint arXiv:2406.06462}, 2024.

\end{thebibliography}
